\PassOptionsToPackage{prologue,dvipsnames,table}{xcolor}
\documentclass[10pt,twocolumn,letterpaper]{article}

\usepackage[pagenumbers]{cvpr} 

\usepackage{graphicx}
\usepackage{amsmath}
\usepackage{amssymb}
\usepackage{booktabs}
\usepackage{multirow}
\usepackage{colortbl}
\usepackage{adjustbox}
\usepackage[dvipsnames]{xcolor}
\usepackage[table]{xcolor}

\definecolor{cvprblue}{rgb}{0.21,0.49,0.74}
\usepackage[pagebackref,breaklinks,colorlinks,allcolors=cvprblue]{hyperref}

\def\paperID{6067} 
\def\confName{CVPR}
\def\confYear{2025}

\title{Empowering Sparse-Input Neural Radiance Fields with Dual-Level \\ Semantic Guidance from Dense Novel Views}

\author{
\begin{tabular}[t]{@{}c@{}}
Yingji Zhong$^1$ \quad Kaichen Zhou$^2$ \quad Zhihao Li$^2$ \quad Lanqing Hong$^2$ \quad Zhenguo Li$^2$ \quad Dan Xu$^1$
\end{tabular}\\[1ex]
\begin{tabular}[t]{@{}c@{}}
$^1$The Hong Kong University of Science and Technology\quad $^2$Huawei Noah's Ark Lab
\end{tabular}\\[0.5ex]
}

\begin{document}
\maketitle

\begin{abstract}
Neural Radiance Fields (NeRF) have shown remarkable capabilities for photorealistic novel view synthesis. One major deficiency of NeRF is that dense inputs are typically required, and the rendering quality will drop drastically given sparse inputs.  In this paper, we highlight the effectiveness of rendered semantics from dense novel views, and show that rendered semantics can be treated as a more robust form of augmented data than rendered RGB. Our method enhances NeRF's performance by incorporating guidance derived from the rendered semantics. The rendered semantic guidance encompasses two levels: the supervision level and the feature level. The supervision-level guidance incorporates a bi-directional verification module that decides the validity of each rendered semantic label, while the feature-level guidance integrates a learnable codebook that encodes semantic-aware information, which is queried by each point via the attention mechanism to obtain semantic-relevant predictions. The overall semantic guidance is embedded into a self-improved pipeline. We also introduce a more challenging sparse-input indoor benchmark, where the number of inputs is limited to as few as 6. Experiments demonstrate the effectiveness of our method and it exhibits superior performance compared to existing approaches.
\end{abstract}    

\vspace{-4mm}
\section{Introduction}
\label{sec:intro}

\begin{figure}[t]
\centering
\includegraphics[width=0.99\linewidth]{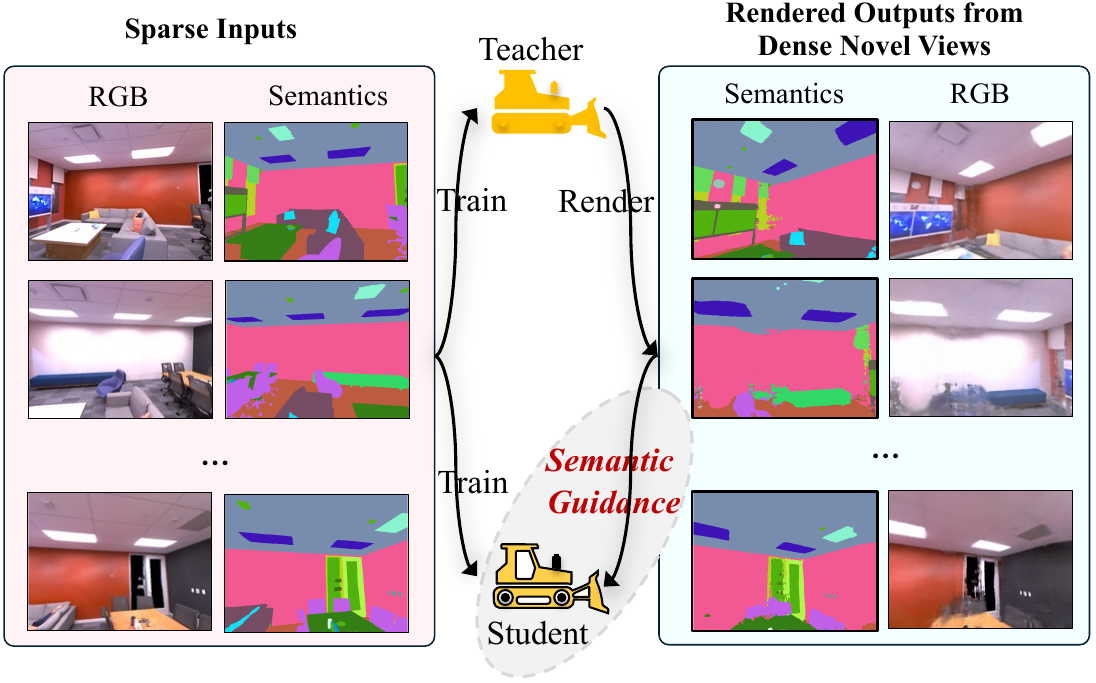}
\vspace{-2mm}
\caption{
Our method exploits rendered semantics from dense novel views to boost the performance of NeRF from sparse inputs. The rendered semantics is fully exploited by semantic guidance, which encompasses supervision-level and feature-level. Our method is embedded into a self-improved framework. }\vspace{-4mm}
\label{fig:teaser}
\end{figure}

Neural implicit representations~\cite{park2019deepsdf,mescheder2019occupancy,sitzmann2019scene,niemeyer2020differentiable,mildenhall2020nerf} have shown superior capabilities in modeling 3D geometry and appearance. Among them, Neural Radiance Fields (NeRF)~\cite{mildenhall2020nerf,barron2021mip,verbin2022ref,liu2020neural,barron2022mip,muller2022instant,barron2023zip} have shown remarkable improvement on novel view synthesis. Albeit impressive, dense input views are required to train the NeRF.
Given sparse input views, it is likely for the NeRF to learn a trivial solution that can explain all training views but fail to model accurate geometry and appearance of the scene, known as the shape-radiance ambiguity~\cite{zhang2020nerf++}. The ambiguity is caused by the incapability of the model to build up correspondences across training views from sparse RGB supervision, causing severe artifacts in novel view synthesis. 

Previous works applied different methods to improve the performance in sparse-input setting~\cite{deng2022depth,kim2022infonerf,niemeyer2022regnerf,jain2021putting,truong2023sparf,wynn2023diffusionerf}. 
Although great improvements have been achieved, the above methods typically overlook the fact that, for a NeRF trained from sparse inputs, its novel-view rendered images can be treated as augmented data. Since the camera poses of novel views are arbitrary, 
the augmented data can be dense. 
Another important observation is that, despite the existing artifacts in rendered novel view images, we can tell the semantics of each part in the image. It also holds for a trained NeRF, i.e., it can render basically accurate semantics in novel views, as shown in Fig.~\ref{fig:teaser}. 
The rich semantics in the dense novel views can be used as augmented data to train another NeRF, 
which can facilitate the model to overcome the ambiguity of the view correspondences, thus leading to improvements in novel view synthesis.

In this paper, we embed the aforementioned motivation into a self-improved framework, dubbed as Dense \textbf{S}emantic Guidance for Neural Radiance Fields from \textbf{S}parse Inputs with \textbf{S}elf-Improvement (S$^3$NeRF). 
As shown in Fig.~\ref{fig:teaser}, after the teacher NeRF has been trained with sparse inputs of RGB and semantics~\cite{zhi2021place}, we can obtain densely rendered outputs of novel views, including rendered RGB and semantics. Along with sparse inputs, we further train the student NeRF with the \emph{rendered semantic guidance}. 
Note that we do not claim the self-improved framework as our main contribution, which is already explored in~\cite{bai2023self,jung2023self}. 
Our main contribution lies in \emph{the first time to explore rendered semantics}, rather than considering rendered RGB as in \cite{bai2023self,jung2023self}, 
and accordingly design different guidance methods to improve the effectiveness of using the rendered semantics. 
Applying rendered RGB as the augmented data might hamper the NeRF training due to artifacts in rendered images, while the rendered semantics can intuitively serve as more reliable augmented data, as we can clearly observe the rendered semantics of different regions regardless of the pixel artifacts of shifted colors or blurs, as illustrated in Fig.~\ref{fig:teaser}. 

\begin{figure*}[th]
\centering
\includegraphics[width=0.85\linewidth]{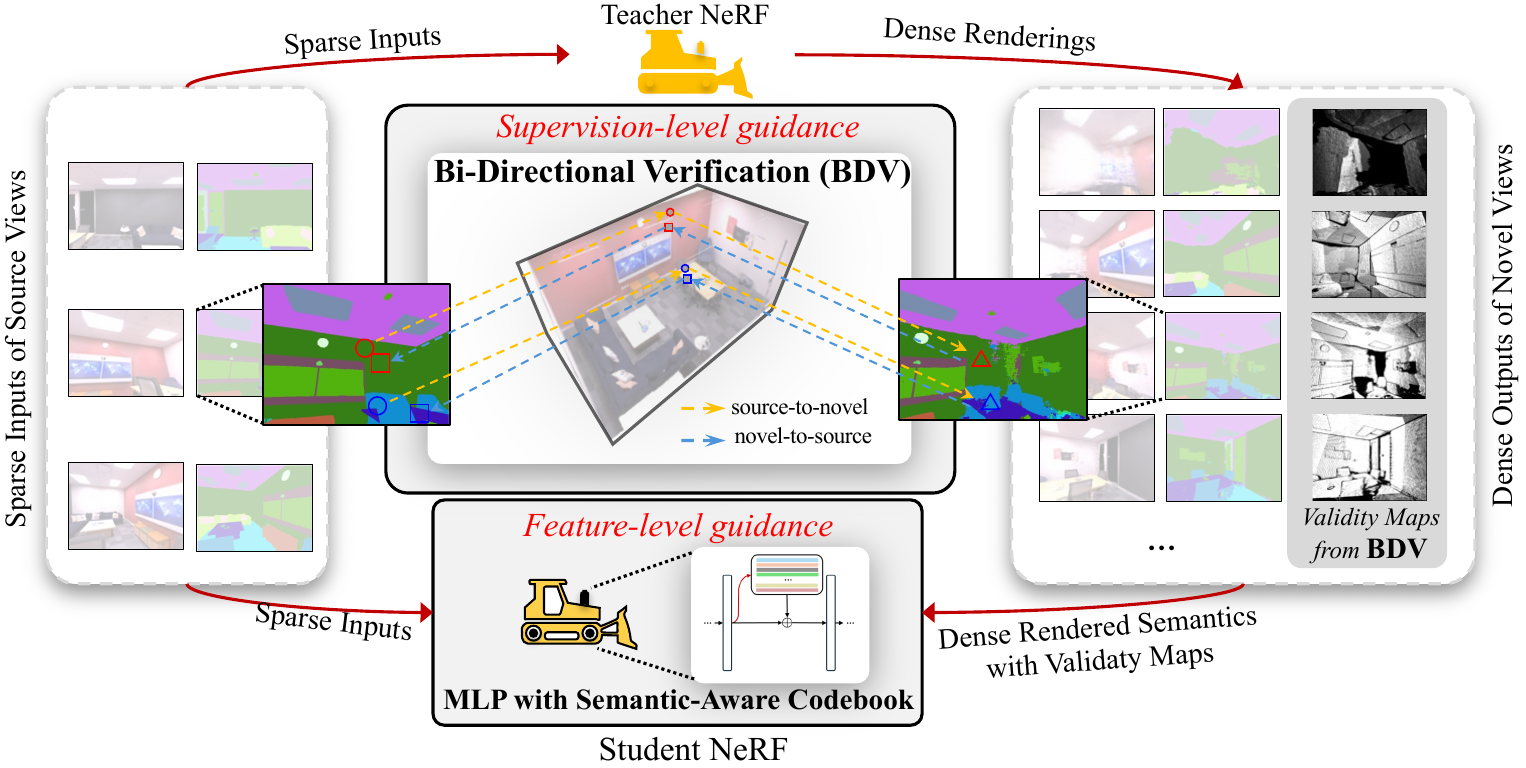}\vspace{-2mm}
\caption{Overview of the proposed S$^3$NeRF, which is built upon a self-improved pipeline. S$^3$NeRF exploits the rendered semantics from the teacher NeRF by two levels of guidance: supervision-level guidance with Bi-Directional Verification (BDV), and feature-level guidance with semantic-aware codebook. 
BDV returns validity maps for each semantic map, indicating the correctness of semantic labels for robust supervision. The semantic-aware codebook encodes correlation among densities, colors, and semantics, further exploiting the underlying information embedded in the semantic labels. The codebook is integrated into the MLP of the student NeRF. 
}\vspace{-5mm}
\label{fig:method_framework}
\end{figure*}

We fully exploit the rendered novel-view semantics as guidance for learning the student NeRF from two perspectives. One is a \emph{supervision-level guidance} and the other is a \emph{feature-level guidance}, as illustrated in Fig.~\ref{fig:method_framework}. Specifically,~\textbf{(i)} for the supervision-level guidance, the rendered semantics serves as an additional supervision signal for the student NeRF. It helps the NeRF to alleviate the ambiguity by building up the correspondences across views with the assistance of semantic labels. 
However, it is unavoidable that there exist incorrect rendered labels, which may harm the learning of correspondences.
Therefore, we propose a Bi-Directional Verification module to tackle the issue of incorrect semantic labels. The rendered semantic label of a pixel in the novel view is considered as valid only if a consensus constraint based on projection is satisfied. 
Hence, the student NeRF is trained to be a decent semantic field, but with limited constraints on the colors in novel views. \textbf{(ii)} To further exploit the supervision from semantic labels, we propose a feature-level guidance module that incorporates a learnable codebook in the MLP. The codebook learns semantic-aware patterns, which are expected to encode the correlation among semantics, colors, and densities. 
For each 3D point that is input into the MLP, its implicit feature queries the codebook via an attention process, which guides the learning of semantic-correlated implicit features for better predictions of colors and densities. 
This module further compensates at the feature level for 3D points that receive only the semantic supervision. 

Previous works on sparse-input NeRF mainly target at face-forwarding scenarios~\cite{jensen2014large,mildenhall2019local}, or 360$^\circ$ scenarios~\cite{mildenhall2020nerf}. In these scenarios, views are sampled in an ``outside-in'' or ``face-forwarding'' manner, meaning that a certain amount of overlap exists across views. Roessle~\textit{et al.}~\cite{roessle2022dense} introduce a sparse-input setting on ScanNet~\cite{dai2017scannet}, where the viewing direction follows an ``inside-out'' pattern, while the number of input views is 18. 
To validate the effectiveness of our method in a more challenging scenario, we introduce a setting based on the indoor scenes of ScanNet$++$~\cite{yeshwanth2023scannet++} and Replica~\cite{straub2019replica}, where the viewing direction is also ``inside-out'', but we largely reduce the number of input views to 6. It is sufficient to cover the entire scene but with remarkably less overlap across views. 


In summary, our contributions are as follows: \textbf{(i)} 
We introduce S$^3$NeRF, the first work leveraging rendered semantics from dense novel views for sparse-input NeRF, which is built upon a self-improved framework; 
\textbf{(ii)} The semantic guidance is explored and incorporated from two perspectives, i.e., supervision-level and feature-level. The former is implemented with a Bi-Directional Verification module, while the latter is realized by a learnable semantic-aware codebook, both of which show strong improvements over the baseline.~The proposed S$^3$NeRF also exhibits competitive performance compared to other works;
\textbf{(iii)} We present an additional sparse-input indoor benchmark utilizing as few as 6 input views, which is significantly more challenging compared to current sparse-input settings.

\section{Related Work}
\label{sec:related}

\noindent \textbf{NeRF from Sparse Inputs.}
Although recent works~\cite{mildenhall2020nerf,barron2021mip,verbin2022ref,liu2020neural,barron2022mip,yan2024gs,muller2022instant,barron2023zip,zhenxing2022switch} have achieved impressive results on novel view synthesis, dense inputs are typically required. 
Some recent works learn a generalizable NeRF by pre-training it on multi-view datasets. Given sparse inputs on the target scene, the pre-trained model performs view synthesis directly or is further finetuned on the sparse inputs~\cite{yu2021pixelnerf,wang2021ibrnet,liu2022neural,chen2021mvsnerf}. 
In contrast, other works train a scene-specific NeRF, utilizing the hand-crafted constraints~\cite{kim2022infonerf,deng2022depth,roessle2022dense,wang2023sparsenerf,zhong2024cvt} or pre-trained priors~\cite{radford2021learning,wynn2023diffusionerf,truong2021learning} to regularize the training. 
In this paper, we also train a scene-specific NeRF from sparse inputs. 
Different with existing works, we utilize the rendered semantics from a trained NeRF as effective augmented data to guide another NeRF in a self-improved manner.

\noindent \textbf{NeRF for 3D Scene Understanding. }Recent works have been exploring to utilize NeRF for 3D scene understanding. 
Zhi~\textit{et al.}~\cite{zhi2021place} propose an innovative framework that jointly encodes the geometry, appearance, and semantic information of a 3D scene into a NeRF. 
Similar encoding processes for scene understanding are utilized in~\cite{vora2021nesf,zhang2023beyond,kundu2022panoptic,siddiqui2023panoptic}. 
These methods target 3D scene understanding with NeRF from dense-view inputs. However, we focus on learning a NeRF from sparse inputs. We also encode semantics into the NeRF, but with a clearly different purpose, which is, we use the semantics to guide the NeRF to improve the geometry and appearance modeling in the sparse-input setting. 

\noindent \textbf{Self-Training NeRF. }Self-training originates from the research field of semi-supervised learning. Self-training methods firstly train a teacher model with sparsely labeled data. The teacher model is then used to generate pseudo labels for unlabeled data, which is combined with the labeled data to train a student model~\cite{zou2019confidence,xie2020self,kirillov2023segment}. There are also attempts of employing the self-training paradigram on NeRF~\cite{bai2023self,jung2023self}, which apply rendered RGB to train the student NeRF. 
However, the rendered RGB values are mostly unreliable, especially in the sparse-input settings where floating artifacts, blurs, and color shiftings widely exist. Since NeRF is trained by RGB correspondences across views, the unreliable rendered RGB values might even exacerbate the ambiguity and thus hamper the modeling performance. 
In this paper, we instead utilize rendered semantics from a trained NeRF as the pseudo labels. We show in the experiments that the semantic guidance is more effective than RGB values, in which it exhibits high semantic label accuracy and superior synthesis performance. 
\vspace{-1mm}
\vspace{-5pt}
\section{The Proposed Method}
In the following, we present the details of our S$^3$NeRF for sparse-input modeling. The overview of our method is illustrated in Fig.~\ref{fig:method_framework}. 
In Sec.~\ref{sec:method_self}, we introduce the self-improved framework, and formulate the learning objective with dense semantic guidance, which is rendered from dense novel views by the teacher NeRF. The semantic guidance is designed to contain a supervision-level (Sec.~\ref{sec:method_supp}) and a feature-level (Sec.~\ref{sec:method_feat}) guidance. We will detail them after the brief review in Sec.~\ref{sec:prelim}. 

\subsection{Preliminary}\label{sec:prelim}
NeRF represents a scene with an MLP network. It predicts the color $\mathbf{c}$ and density $\sigma$ of each point $\mathbf{x} \in \mathbb{R}^3$ in the 3D space, and the color prediction is additionally based on a 2D viewing direction $\mathbf{d} \in \mathbb{R}^2$. The coordinates of each point and the viewing direction are transformed by the positional encoding $\gamma(\cdot)$ before inputting into the MLP. The network prediction can be briefly expressed as: 
$(\mathbf{c}, \sigma)=\rm{MLP}(\gamma(\mathbf{x}), \gamma(\mathbf{d}))$. 
NeRF renders a pixel by casting a ray through the scene and a certain number of points along the ray are sampled, whose densities and colors are predicted by the MLP. The color of a pixel is obtained by volume rendering, which aggregates the colors of all sampled points based on their densities, formulated as: 
$C(\textbf r)=\sum_{i=1}^N T_i(1-\rm exp(-\sigma_{i}\delta_{i}))\mathbf{c}_i$, 
where $\delta_i$ refers to the distance between two adjacent samples and $T_i$ is the accumulated transmittance calculated by: $T_i=\rm exp(-\sum_{j=1}^{i-1}\sigma_j \delta_j)$. With the rendered colors $C(\mathbf{r})$ and ground truth colors $C_{\rm gt}(\mathbf r)$, MLP is trained with a reconstruction loss: 
\begin{equation}\label{eq:recon_loss}
\mathcal{L}_{\rm{recon}} = \frac{1}{\left| \mathbf{r} \right|}\sum_{\textbf{r}}\Vert C(\mathbf r) - C_{\rm gt}(\mathbf r) \Vert ^ 2. 
\end{equation}

\noindent \textbf{Semantic NeRF. }Zhi~\textit{et al.}~\cite{zhi2021place} extends the above framework by encoding semantics into the NeRF. Beside color $\mathbf{c}$ and density $\sigma$, they use the MLP to predict semantic logits $\mathbf{g}$, as illustrated in Fig.~\ref{fig:method_feat} (a). 
Volume rendering is then applied on the logits of each point along the ray: $G(\textbf r)=\sum_{i=1}^N T_i(1-\rm exp(-\sigma_{i}\delta_{i}))\mathbf{g}_i$. The MLP is then trained with the $\mathcal{L}_{\rm{recon}}$ and the following semantic loss $\mathcal{L}_{sem}$: 
\begin{equation}\label{eq:sem_loss}
\mathcal{L}_{\rm{sem}} = \frac{1}{\left| \mathbf{r} \right|}\sum_{\textbf{r}} {\rm{CE}}({\rm{softmax}}(G(\mathbf r)), S_{\rm gt}(\mathbf r)), 
\end{equation}
where ${\rm{softmax}}(\cdot)$ transforms the logits into a distribution, ${\rm{CE}}(\cdot)$ and $S_{\rm gt} (\mathbf r)$ refers to the cross-entropy loss and the ground truth semantic label, respectively. This makes the NeRF not only a radiance field but also a semantic field. 
In this paper, we use semantics to guide the modeling of geometry and color for sparse-input NeRF.

\begin{figure}[t]
\centering
\includegraphics[width=0.99\linewidth]{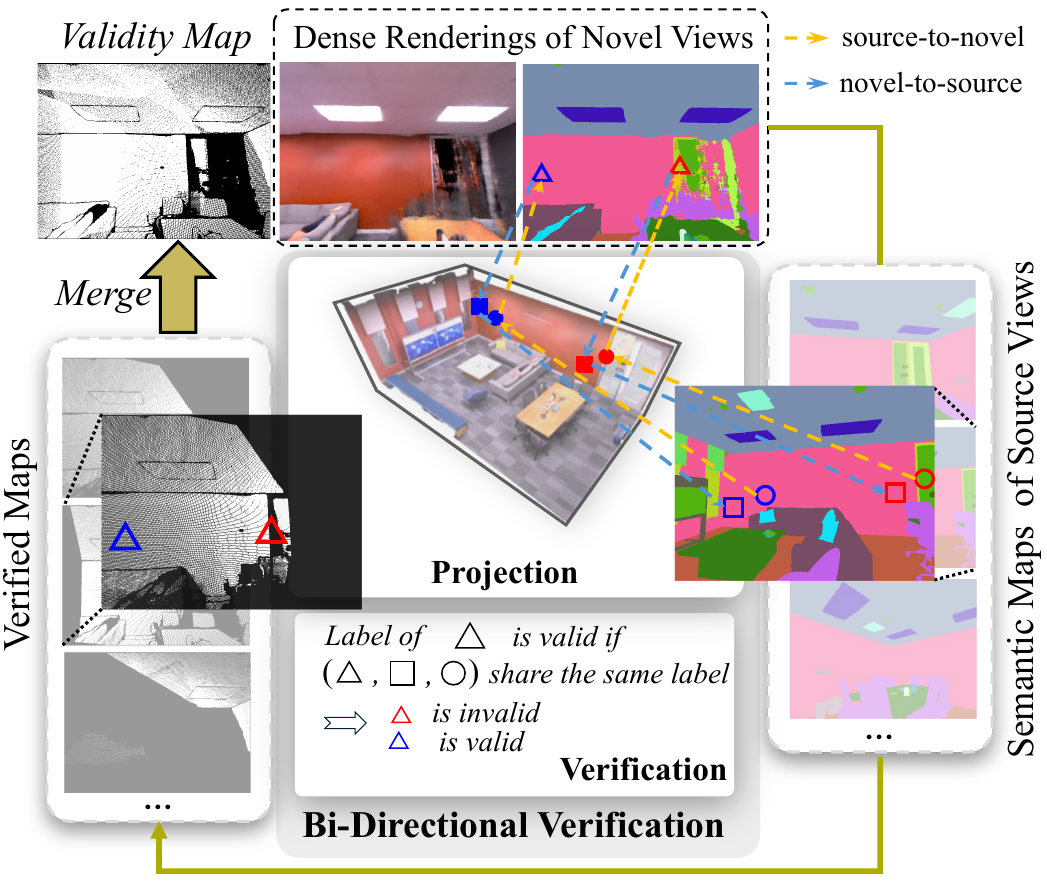}\vspace{-2mm}
\caption{Illustration of the proposed Bi-Directional Verification (BDV) module. For each {rendered} semantic map, BDV is applied between each source view and it by projection and verification. 
Based on results from the projection, the verification step produces a consensus-based verified map.
A validity map is created by merging verified maps from all source views, reflecting the accuracy of the rendered semantic map. Depth maps are omitted. 
}\vspace{-4mm}
\label{fig:method_bdv}
\end{figure}

\subsection{Self-Improved NeRF with Rendered Semantics from Dense Novel Views}\label{sec:method_self}
\vspace{-0pt}
The self-improved framework applied in this work involves training a teacher and a student NeRF, respectively. The outputs of novel views rendered from the teacher NeRF are treated as the augmented data to train the student NeRF. 

Given sparse inputs of $N$ calibrated images as well as their corresponding semantics, i.e., $\{\mathbf{I}_i, \Phi_i, \mathbf{S}_i\}_{i=1}^N$, where three values refer to RGB values, the camera pose, and semantic labels of each image, we firstly train the teacher NeRF with the reconstruction loss of Eq.~\ref{eq:recon_loss} and the semantic loss of Eq.~\ref{eq:sem_loss}. 
We observe that limited semantics of the sparse inputs degrade the performance of novel view synthesis. Thus, we detach the semantic branch of Fig.~\ref{fig:method_feat} (a) from the main branch. After training, the teacher NeRF can render images as well as their semantic maps from novel views. The rendered outputs are represented by $\{\hat{\mathbf I}_j, \hat{\Phi}_j, \hat{\mathbf S}_j, \hat{\mathbf D}_j\}_{j=1}^T$, where $T$ refers to the number of novel views, which can be typically larger than $N$, and $\{\hat{\mathbf{D}}_j\}_{j=1}^T$ refers to the rendered depth maps. 

Previous works~\cite{bai2023self,jung2023self} attempt to exploit the rendered RGB values $\{\hat{\mathbf I}_j\}_{j=1}^T$ as the augmented data, with which the student NeRF is further trained. However, the rendered RGB values are not reliable due to the existence of color shiftings and blurry regions in the rendered images. 

Instead, we propose to use the rendered semantics $\{\hat{\mathbf S}_j\}_{j=1}^T$ as the augmented data. The rendered semantic labels exhibit greater robustness, as the trained teacher NeRF can render accurate semantic labels for most regions, even for those blurry ones. 
To utilize the rendered semantics, a straightforward way is to supervise the rays of novel views with their rendered semantic labels by the semantic loss $\mathcal{L}_{\rm{sem}}$. Although the rendered semantic labels show higher robustness, there still exists misclassified labels that might hamper correspondence learning during the training.
We thus introduce a weighting factor to adjust the impact of the incorrect labels. The training objective of the student NeRF is then formulated as: 
\begin{equation}\label{eq:form}
\setlength{\abovedisplayskip}{0.05cm}
\setlength{\belowdisplayskip}{0.05cm}
\begin{aligned}
    \mathcal{L} = \mathcal{L}_{\rm{recon}} &+ \frac{\lambda_{\rm{sem}}}{\left| \mathbf{r} \right|+\left| \hat{\mathbf{r}} \right|} \cdot \left(\sum_{\mathbf{r} \in \mathcal{R}} {\rm{CE}}({\rm{softmax}}(G(\mathbf{r})), S_{\rm{gt}}(\mathbf{r})) \right. \\[-1.5ex]
    &\left. + \sum_{\hat{\mathbf{r}} \in \hat{\mathcal{R}}} w(\hat{\mathbf{r}})\cdot{\rm{CE}}({\rm{softmax}}(G(\hat{\mathbf{r}})), \hat{S}_{\rm{gt}}(\hat{\mathbf{r}})) \right),
\end{aligned}
\end{equation}
where the semantic guidance is balanced by the $\lambda_{\rm{sem}}$ term. 
$\mathcal{R}$ and $\hat{\mathcal{R}}$ represent the sets of rays from sparse input views and novel views, respectively; $\hat{\mathbf{r}}$ refers to sampled rays from rendered novel views. 
$w(\hat{\mathbf{r}})$ and $\hat{S}_{\rm{gt}}(\hat{\mathbf{r}})$ represent the weights and pseudo semantic labels for the rays, respectively. In the following, we elaborate on the detailed strategy of deciding $w(\hat{\mathbf r})$ for the rays.

\vspace{-0mm}
\subsection{Supervision-level Guidance}\label{sec:method_supp}
\vspace{-0mm}
In this section, we propose a Bi-Directional Verification (BDV) module to determine the validity of each rendered semantic label. Thus, the weighting factor $w(\hat{\mathbf{r}})$ in Eq.~\ref{eq:form} can be either 1 or 0, indicating whether the label is valid or invalid. 
Subsequently, the term ``source views'' is utilized to refer to the sparse inputs. 

As shown in Fig.~\ref{fig:method_framework}, BDV is applied between each source view and each rendered novel view. 
Fig.~\ref{fig:method_bdv} presents more details about the BDV module. 
Concretely, the inputs into the BDV module consist of $\{{\Phi}, {\mathbf S}, {\mathbf D}\}$ and $\{\hat{\Phi}, \hat{\mathbf S}, \hat{\mathbf D}\}$, where each of the symbols in the triplet refers to the camera pose, the semantic map and the depth map, respectively. Note that $\mathbf D$ is also rendered by the teacher NeRF. For clarity in the following statements, we drop the subscripts to consider one source view and one novel view. 

The motivation behind BDV is straightforward: if a rendered semantic label is accurate, we should be able to identify the same semantic label in corresponding regions of at least one specific source view. 
BDV consists of two key steps: projection and verification. The projection step focuses on establishing correspondences between regions in the source view and the novel view. Subsequently, the verification step determines the validity of each semantic label in the novel view, by leveraging the information obtained during the projection step.

\begin{figure}[t]
\centering
\includegraphics[width=0.99\linewidth]{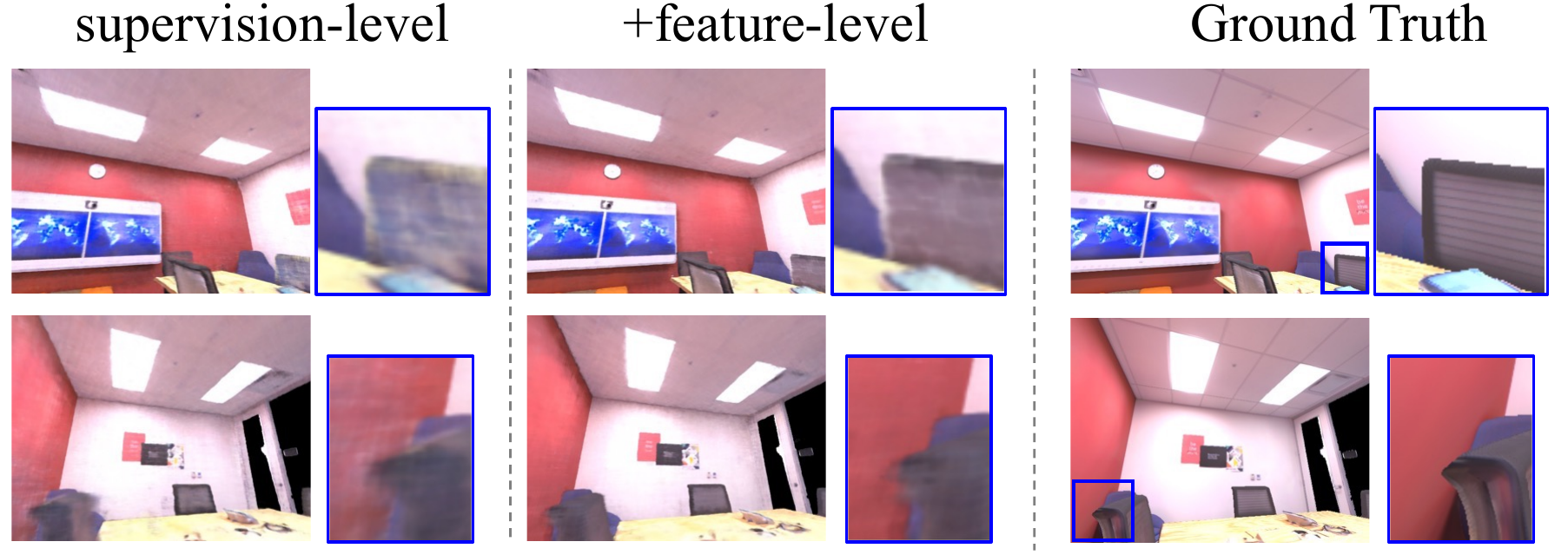}
\vspace{-3mm}
\caption{
    Improved color predictions by the feature-level guidance, leveraging semantic-relevant information in the codebook. }\vspace{-4mm}
\label{fig:vis_sec}
\end{figure}

\noindent \textbf{Projection. }
Given the rendered depth maps $\mathbf D$ and $\hat{\mathbf D}$, we employ projection to establish correspondences between regions of the source view and the novel view. Our objective is to determine the validity of each semantic label in the novel view, by identifying its corresponding region in the source view, and verifying whether it matches the semantic label assigned to that source view region. While the depth maps may not provide accurate pixel-to-pixel correspondences, we have observed that they are generally precise enough to establish pixel-to-region correspondences in most cases, as depicted in Fig.~\ref{fig:method_bdv}. To enhance the robustness in determining the correctness of the semantic labels, our proposed BDV employs two directions of projection: source-to-novel and novel-to-source projections. These projections serve as the basis for the verification step.

\noindent \textit{Source-to-novel projection: }Given the camera poses $\Phi$ and $\hat{\Phi}$, along with the rendered depth maps $\mathbf D$, for the coordinates of each pixel $\mathbf p$ in the source view, we can project it onto the novel view, and obtain the projected coordinates $\hat{\mathbf p}_{\rm{src} \rightarrow \rm{nov}}$ using the following formula: 
\begin{equation}\label{eq:src2nov}
\setlength{\abovedisplayskip}{0.2cm}
\setlength{\belowdisplayskip}{0.2cm}
\hat{\mathbf p}_{\rm{src} \rightarrow \rm{nov}} \sim K T_{\Phi \rightarrow \hat{\Phi}} \mathbf{D}(\mathbf p) K^{-1} \mathbf{p}, 
\end{equation}
where $K$ refers to the camera intrinsic matrix, and $T_{\Phi \rightarrow \hat{\Phi}}$ represents the pose transition from the source view pose to the novel view pose. $\mathbf p$ is input to Eq.~\ref{eq:src2nov} in the form of homogeneous coordinates but we omit it for simplicity. 

Intuitively, if $\hat{\mathbf p}_{\rm{src} \rightarrow \rm{nov}}$ in the novel view is located at the same region with $\mathbf p$ in the source view, the correctness of the semantic label can be determined by checking if $\hat{\mathbf S}(\hat{\mathbf p}_{\rm{src} \rightarrow \rm{nov}})$ is identical to $\mathbf S(\mathbf p)$. However, the depth map $\mathbf D$ of the source view may contain large errors in some regions, which will result in wrong projections. The semantic label may still be incorrect even if it passes the above checking. Therefore, we also include the rendered depth of novel views to comprehensively determine the correctness of the semantic labels, by utilizing the inverse projection, i.e., \textit{Novel-to-source projection}: 
\begin{equation}
\setlength{\abovedisplayskip}{0.2cm}
\setlength{\belowdisplayskip}{0.2cm}
{\mathbf p}_{\rm{nov} \rightarrow \rm{src}} \sim K T_{\hat{\Phi} \rightarrow \Phi} \hat{\mathbf{D}}(\hat{\mathbf p}_{\rm{src}\rightarrow\rm{nov}}) K^{-1} \hat{\mathbf p}_{\rm{src}\rightarrow\rm{nov}}.
\end{equation}


\noindent \textbf{Verification. }With the bi-level projection, we have a projection chain that composes of a triplet, i.e., $(\mathbf p, \hat{\mathbf p}_{\rm{src} \rightarrow \rm{nov}}, {\mathbf p}_{\rm{nov} \rightarrow \rm{src}})$, which is respectively represented by the circle, triangle, and square in Fig.~\ref{fig:method_bdv}. 
To obtain the verified map $\hat{\mathbf M}$ which verifies the correctness of the semantic labels of the novel view $\hat{\mathbf S}$, we apply the simplest verification step based on consensus: 
\begin{equation}\label{eq:ver}
\setlength{\abovedisplayskip}{0.2cm}
\setlength{\belowdisplayskip}{0.2cm}
    \hat{\mathbf M} =
    \begin{cases}
        1 & \tiny{\rm{if}\quad \mathbf{S}(\mathbf{p})=\hat{\mathbf S}(\hat{\mathbf p}_{\rm{src} \rightarrow \rm{nov}})= \mathbf{S}({\mathbf p}_{\rm{nov} \rightarrow \rm{src}})}\\
        0 & \rm{otherwise}
    \end{cases}, 
\end{equation}
which means that the semantic label in the novel view is considered valid only if its assigned triplet shares the same label. 
Note that we only consider the coordinates on the novel view projected from the source view. A simple illustration of the verified map can be found in Fig.~\ref{fig:method_framework}. 

\noindent \textit{Merge Verified Maps into a Validity Map: }The verified map of Eq.~\ref{eq:ver} is obtained between one source view and one novel view. However, the correctness of the semantic labels in the novel view should be verified after checking with all $N$ source views. Therefore, for each semantic map of the novel view, we repeat the above projection and verification steps with other source views, and obtain totally $N$ verified maps $\{\hat{\mathbf M}_i\}_{i=1}^N$. 
We can then obtain the validity map $\hat{\mathbf V}$ for the novel view by: 
$\hat{\mathbf V} = \bigcup_{i=1}^{N} \hat{\mathbf M}_i, $
where $\bigcup$ denotes the ``element-wise or'' operation. 
As shown in Fig.~\ref{fig:method_framework} and Fig.~\ref{fig:validity}, the validity maps can reflect the correctness of semantic labels. 
Regarding Eq.~\ref{eq:form}, we can set $w(\hat{\mathbf r})$ to 0 or 1 according to its value in the validity map. 

\begin{figure*}[t]
  \centering
  \includegraphics[width=0.99\linewidth]{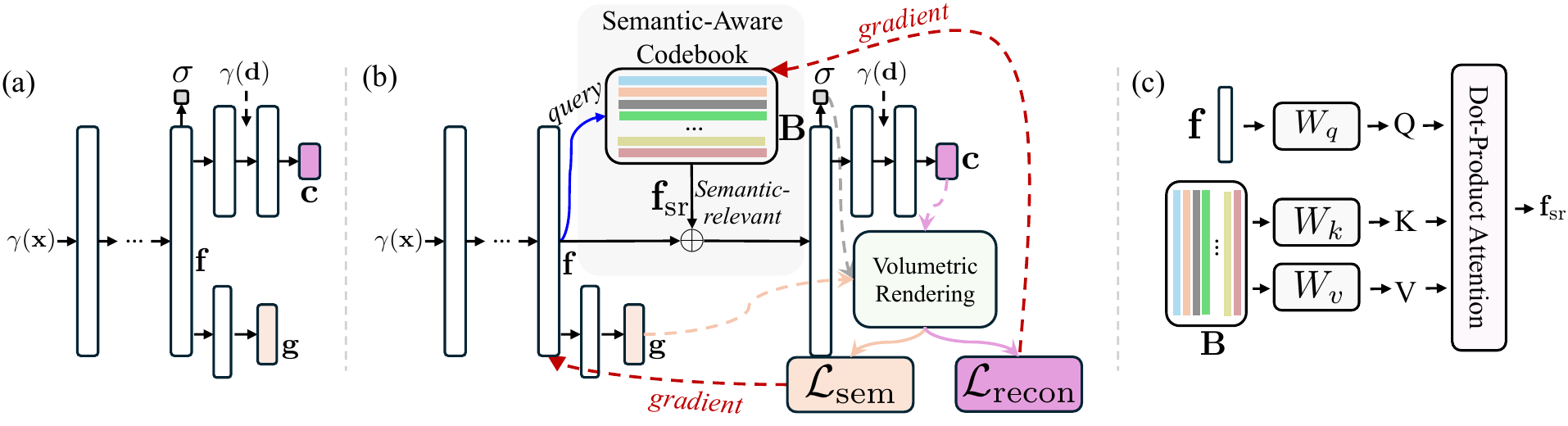}\vspace{-3mm}
  \caption{\textbf{(a)} MLP of the semantic NeRF~\cite{zhi2021place}, where $\mathbf c$, $\sigma$ and $\mathbf g$ refer to color, density, and semantic logits of each input 3D point. $\mathbf f$ is the implicit feature used to predict the density. \textbf{(b)} Our proposed MLP for feature-level guidance incorporates a semantic-aware codebook to encode the correlation among densities, colors, and semantics. Each 3D point queries the codebook via an attention in \textbf{(c)}. The codebook $\mathbf B$ is updated by the gradient from the reconstruction loss, while the semantic field is learned from the semantic loss.  
  }
  \vspace{-4mm}\label{fig:method_feat}
\end{figure*}

\vspace{-0mm}
\subsection{Feature-level Guidance}\label{sec:method_feat}
\vspace{-0pt}
Eq.~\ref{eq:form} with the proposed supervision-level guidance can alleviate the ambiguity problem in the sparse-input setting by using the rendered semantic labels to build up correspondences across views. As mentioned in Sec.~\ref{sec:method_self}, the rendered RGB cannot be considered as reliable augmented data and we do not use them for supervision. This makes the trained student NeRF a decent semantic field, 
exhibiting improved rendering of multi-view consistent semantics, but with limited enhancement in view synthesis (i.e., appearance) due to the absence of RGB supervision for augmented rays. 

We observe that the semantics are correlated to the color, e.g., walls in a scene only exhibit limited colors. Moreover, the encoded semantics of 3D points in the semantic field are even related to their densities. For instance, for a point encoded with certain semantics, it density can be high. Therefore, the rendered semantics not only provide semantic labels for the student NeRF, enabling it to function as a reliable semantic field, but also convey the inner semantic-relevant information, such as colors, which is not employed by the supervision-level guidance. 
By leveraging the correlation between semantics and colors, we can model a more accurate radiance distribution of the scene with the guidance from the correlation. 
Furthermore, it can also compensate for those novel views that lack RGB supervision to achieve better color predictions, 
thus improving the performance of view synthesis. 

The rendered semantic labels from dense novel views can train the MLP to represent a decent semantic field, using the vanilla structure as depicted in Fig~\ref{fig:method_feat} (a). Thus, the implicit feature $\mathbf f$ is trained to be well-encoded with semantics. 
To further exploit the rendered semantic guidance from the teacher NeRF, 
we propose a feature-level guidance that guides the learning of the implicit feature not only for well-encoded semantics, but also for better predictions of colors and densities. 
To implement this, we incorporate a learnable codebook within the MLP, as illustrated in Fig.~\ref{fig:method_feat} (b). The codebook learns semantic-aware patterns, and captures the correlation among semantics, colors, and densities. 
Consequently, the predictions of the density and color are based on the semantic-relevant information that is queried by the semantics encoded in the feature $\mathbf f$, leading to more accurate predictions. 
We denote the codebook as: $\mathbf{B}\in \mathbb{R}^{K\times d}$, which contains $K$ learnable embeddings of dimension $d$. 

For each 3D point that is input to the MLP, after obtaining the implicit feature $\mathbf f$ through the feed-forwarding, we extract the semantic-relevant information from the codebook with a query operation: 
$\mathbf{f}_{\rm{sr}} = \rm{Query}(\mathbf{f}, \mathbf B)$,  
where $\mathbf f_{\rm{sr}}$ denotes the queried feature. The query operation is implemented with an attention process: 
\begin{equation}\label{eq:attn1}
\setlength{\abovedisplayskip}{0.15cm}
\setlength{\belowdisplayskip}{0.15cm}
\begin{aligned}
    \mathbf{q}=W_q^T \mathbf{f}&, \quad \mathbf{K}=W_k^T \mathbf{B}, \quad \mathbf{V}=W_v^T \mathbf{B}, \\
     \mathbf{f}_{\rm{sr}} &= \sum_{i=1}^K {\rm{softmax}}(\mathbf{q} \mathbf{K}^T)_i {\mathbf{V}}_{i}, 
\end{aligned}
\end{equation}
where $W_q$, $W_k$, and $W_v$ $\in \mathbb{R}^{d \times d}$ represent the learnable weights of transforming the inputs to the query, the key, and the value embeddings for the attention, respectively. 
While $\mathbf{f}_{\rm{sr}}$ contains semantic-relevant information, using it directly for predictions may not yield satisfactory outcomes, due to the inherent heterogeneity within each semantic class. For instance, the walls may show different colors in a scene. 
To address this issue, as depicted in Fig.~\ref{fig:method_feat} (b), we employ element-wise addition between $\mathbf f$ and $\mathbf f_{\rm{sr}}$, combing the position-specific with the semantic-relevant information to obtain a more comprehensive representation, which is used for color and density predictions. 
The codebook can be learned from the gradient of the reconstruction loss (Eq.~\ref{eq:recon_loss}), as illustrated in Fig.~\ref{fig:method_feat} (b). Fig.~\ref{fig:vis_sec} qualitatively demonstrates that, with the feature-level guidance, the student NeRF can render more accurate colors.

\section{Experiments}

\begin{table*}[t]
	\centering
	\huge
        \makebox[0.325\textwidth][l]{
	\begin{subtable}{0.325\textwidth}
		\centering
		\adjustbox{width=0.97\linewidth}{%
                \renewcommand{\arraystretch}{1.15}
			\begin{tabular}{lccc}
				\toprule[2.5pt]
				\textbf{(a)} & PSNR$\uparrow$ & SSIM$\uparrow$ & LPIPS$\downarrow$ \\ \midrule
				baseline & 21.17 & 0.779 & 0.395 \\
				\cellcolor{gray!25}+supervision-level & \cellcolor{gray!25}\textbf{21.69} & \cellcolor{gray!25}\textbf{0.780} & \cellcolor{gray!25}\textbf{0.387} \\
				\quad w/o rendered labels & 20.49 & 0.747 & 0.416 \\
				\quad w/o verification & 21.35 & 0.774 & 0.409 \\
				\bottomrule[2.5pt]
			\end{tabular}
		}
	\end{subtable}
        }
	\hspace{0.0\textwidth} 
	\makebox[0.346\textwidth][c]{
        \begin{subtable}{0.346\textwidth}
		\centering
		\adjustbox{width=\linewidth}{%
                \renewcommand{\arraystretch}{1.15}
			\begin{tabular}{lccc}
				\toprule[2.5pt]
				\textbf{(b)} & PSNR$\uparrow$ & SSIM$\uparrow$ & LPIPS$\downarrow$ \\ \midrule
				baseline+{supervision-level} & 21.69 & 0.780 & 0.387 \\
				\cellcolor{gray!25}{+feature-level} (\textbf{S$^3$NeRF}) & \cellcolor{gray!25}\textbf{22.21} & \cellcolor{gray!25}\textbf{0.787} & \cellcolor{gray!25}\textbf{0.364} \\
				\quad w/o $\mathcal{L}_{sem}$ & 21.69 & 0.783 & 0.376 \\
				\quad w pre-trained codebook & 20.28 & 0.764 & 0.404 \\ 
				\bottomrule[2.5pt]
			\end{tabular}
		}
	\end{subtable}
        }
	\hspace{0.0\textwidth} 
	\makebox[0.28\textwidth][r]{
        \begin{subtable}{0.28\textwidth}
		\centering
		\adjustbox{width=1.01\linewidth}{%
                \renewcommand{\arraystretch}{1.15}
			\begin{tabular}{lccc}
				\toprule[2.5pt]
				\textbf{(c)} & PSNR$\uparrow$ & SSIM$\uparrow$ & LPIPS$\downarrow$ \\ \midrule
				baseline-12view &23.30 & 0.814 & 0.295 \\
				\cellcolor{gray!25}Ours-12view &\cellcolor{gray!25}\textbf{24.00} & \cellcolor{gray!25}\textbf{0.818} & \cellcolor{gray!25}\textbf{0.277} \\
                    baseline-18view &24.96 &0.827 &0.282 \\
                    \cellcolor{gray!25}Ours-18view &\cellcolor{gray!25}\textbf{25.80} &\cellcolor{gray!25}\textbf{0.837} &\cellcolor{gray!25}\textbf{0.253} \\
				\bottomrule[2.5pt]
			\end{tabular}
		}
	\end{subtable}
	}\vspace{-3mm}
	\caption{
    Ablation studies on the \textbf{(a) supervision-level} and  \textbf{(b) feature-level} guidance on the ScanNet$++$ dataset. 
    \textbf{(c)} 
    Effectiveness of our method with more input views on the Replica dataset. }\vspace{-4mm}
	\label{tab:eff_mod}
\end{table*}

\subsection{Experimental Settings}\vspace{-2pt}
\textbf{Datasets. }Our experiments are conducted on two indoor datasets: Replica~\cite{straub2019replica} and ScanNet$++$~\cite{yeshwanth2023scannet++}, using a sparse-input setting with as few as 6 input images for training.
Replica is a synthetic dataset, while ScanNet$++$ is a high-quality realistic dataset. Both datasets provide semantic labels. We select 6 scenes from Replica and 4 scenes from ScanNet$++$ for our experiments. For each scene, the chosen 6 views of sparse inputs cover the entire scene. This setting is more challenging compared to previous sparse-input settings due to its ``inside-out'' viewing direction and limited overlap across views.
We also evaluate our method on LLFF~\cite{mildenhall2019local} and DTU~\cite{jensen2014large}, which are widely-used benchmarks for sparse-input modeling. 

\noindent \textbf{Baseline. }In our experiments, we adopt Mip-NeRF~\cite{barron2022mip} with the monocular depth regularization~\cite{yu2022monosdf} as the baseline, and we validate the effectiveness of our approach on it. 
Accordingly, we treat the baseline method as the teacher NeRF within our self-improved framework, and it remains fixed throughout the subsequent experiments.

\noindent \textbf{Performances. }In the tables below, the reported results regarding our method pertain to the performance of the student NeRF, except the entry labeled ``baseline'', which is the performance of the teacher NeRF.

\vspace{-0mm}
\subsection{Ablation Studies}
Our proposed S$^3$NeRF is a self-improved framework with the rendered semantic guidance from a teacher NeRF (baseline), which is composed of two levels, {i.e.}, supervision-level and feature-level. In the following, we conduct the ablation studies on the ScanNet++ dataset unless mentioned.  

\begin{figure}[t]
  \centering
  \vspace{-1mm}
  \includegraphics[width=0.95\linewidth]{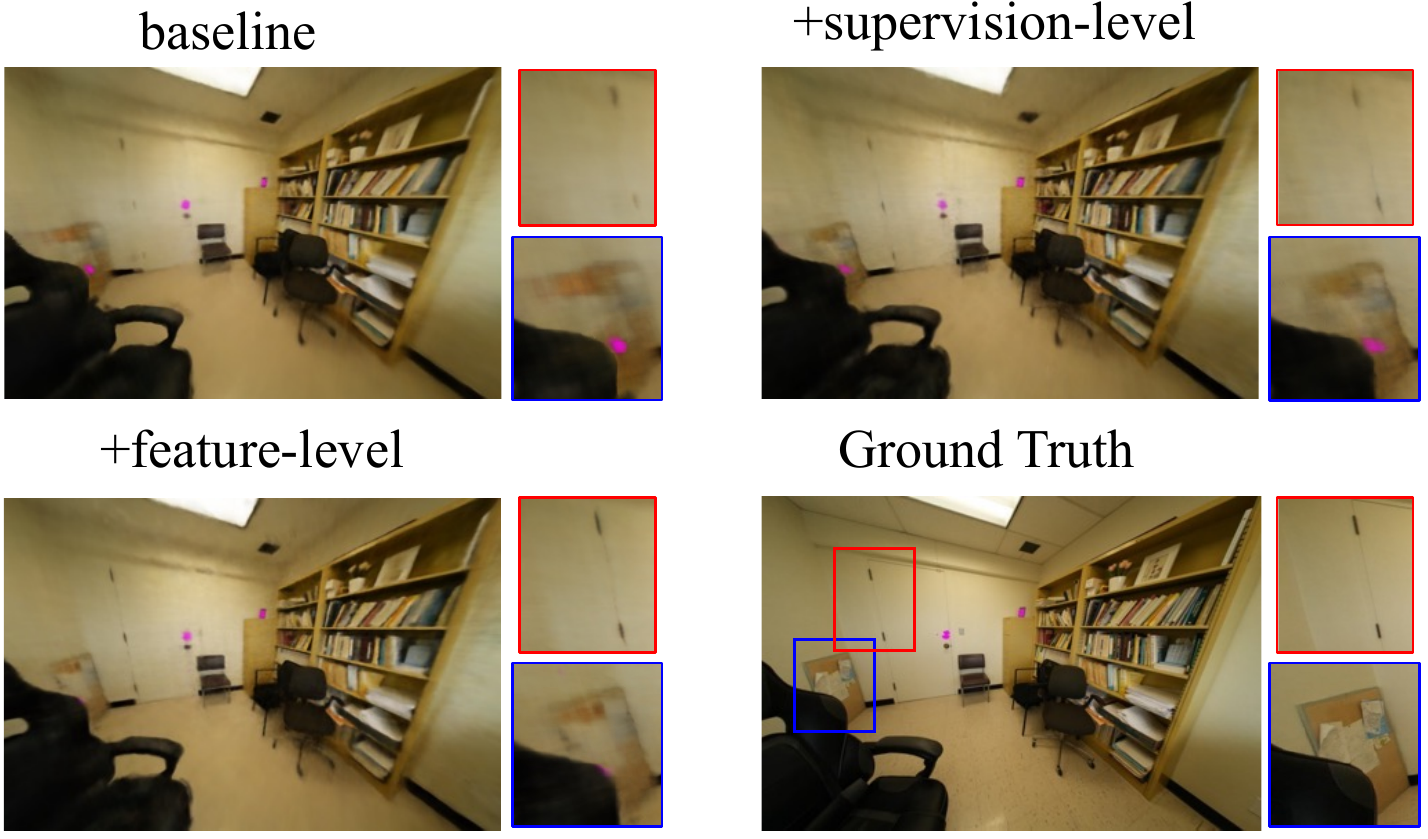}
  \vspace{-3mm}
  \caption{Efficacy of the supervision-level and feature-level guidance. The supervision-level guidance improves the quality in the object boundary regions, while the feature-level guidance boosts the quality by making the model predict more accurate colors. }\vspace{-2mm}
  \label{fig:vis_eff}
\end{figure}

\begin{figure}[t]
  \centering
  \vspace{-0mm}
  \includegraphics[width=0.96\linewidth]{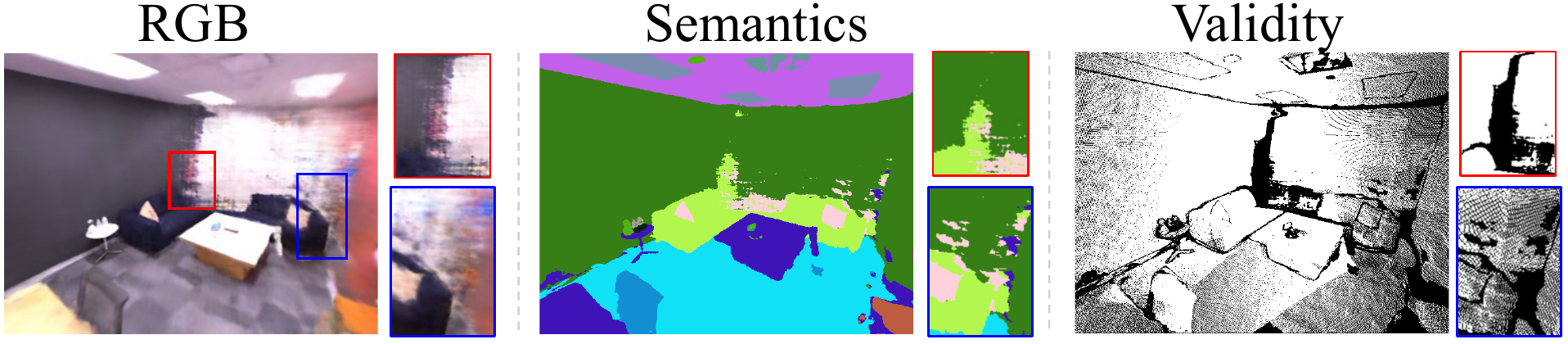}
  \vspace{-3mm}
  \caption{Validity maps from the proposed BDV module. The black regions are recognized as invalid and their enclosed semantic labels are not used in the semantic loss for the student NeRF. }\vspace{-4.5mm}
  \label{fig:validity}
\end{figure}

\noindent \textbf{Effect of the supervision-level guidance. }The supervision-level guidance assists the training of the student NeRF with the rendered semantic labels that are verified by the Bi-Directional Verification (BDV) module. As presented in Tab.~\ref{tab:eff_mod} (a), the supervision-level guidance enhances the performance by more than 0.5 PSNR. Fig.~\ref{fig:vis_eff} demonstrates that the guidance notably increases the quality of the blurry regions around object boundaries. 
However, training with rendered semantic labels without the BDV module may negatively impact the performance, as indicated by the SSIM and LPIPS drops of ``w/o verification'' in Tab.~\ref{tab:eff_mod}.
This is due to the influence of incorrect labels, which exacerbate ambiguity during training and hinder performance. Fig.~\ref{fig:validity} shows that our method can effectively filter out incorrect semantic labels. 
To show that the performance gain is attributed to the semantic labels from the rendered novel views, rather than the ones from the source views, we conduct an experiment where NeRF is trained solely with semantic labels from the source views, denoted as ``w/o rendered labels''. Compared to the baseline PSNR of 21.17, the performance drops to 20.49, indicating that the semantic labels can assist the learning of NeRF only if they are dense enough to build up the correspondences across views. 

\noindent \textbf{Effect of the feature-level guidance. }The feature-level guidance incorporates a semantic-aware codebook into the MLP. In this part, we investigate the impact of this guidance. As indicated in Tab.~\ref{tab:eff_mod} (b), the proposed feature-level guidance brings a PSNR improvement of over 0.5 compared to the model with supervision-level guidance. 
Fig~\ref{fig:vis_sec} and Fig.~\ref{fig:vis_eff} show that, the feature-level guidance improves the synthesis quality, particularly in terms of more accurate color predictions, e.g., the colors of the wall beside the board and the hinges. This is attributed to the correlation between semantics and colors that are encoded in the codebook. 
To study the contribution of the codebook structure, we exclude the semantic loss and denote the variant as ``w/o $\mathcal{L}_{sem}$'' in the table. The structure yields a similar performance gain as the supervision-level guidance. It complements the semantic loss, and combining both results in the largest gain.  
Yin~\textit{et al.}~\cite{yin2022coordinates} also utilize a codebook that is pre-trained from ImageNet. Compared to our proposed Fig.~\ref{fig:method_feat} (b), they extract the visual clues from the codebook at the MLP input, which cannot utilize the feature with encoded semantics. Additionally, the codebook is pre-trained and is thus semantic-agnostic. We also evaluate their structure, denoted as ``w pre-trained codebook'' in the table. Our method outperforms theirs by a large margin, validating the effectiveness of the proposed learnable semantic-aware codebook.



\begin{figure*}[t]
  \centering
  \includegraphics[width=0.98\linewidth]{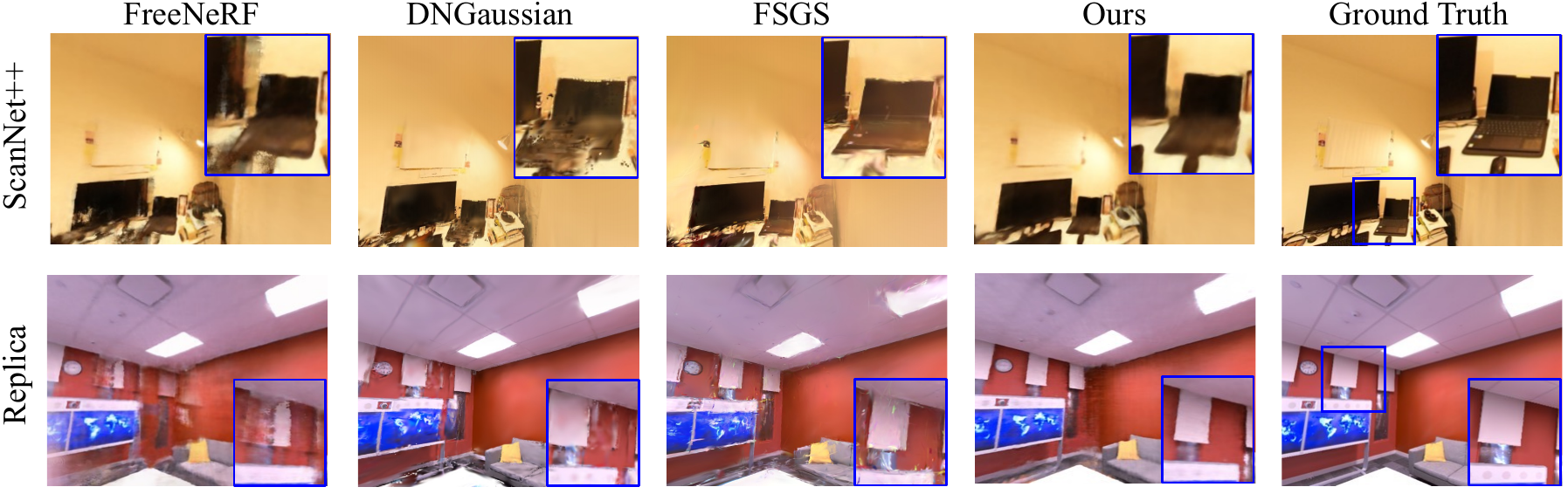}
  \vspace{-3mm}
  \caption{Qualitative comparisons on ScanNet$++$ and Replica datasets. Our method preserves the global structure more effectively, avoiding distortion and severe floaters, while also capturing finer local details such as edges.
  }\vspace{-4mm}
  \label{fig:sota}
\end{figure*}



\noindent \textbf{Performance with more input views. }We also examine the effectiveness of our method as the number of input views increases on the Replica dataset. Table~\ref{tab:eff_mod} (c) shows that our approach consistently improves the performance with more views. Specifically, our method brings PSNR improvement of over 0.5 for input views of both 12 and 18. This indicates that our method does not suffer over-regularization issues.

\begin{table}[t]
\Large
\centering
\vspace{-1mm}
\resizebox{1.\linewidth}{!}{
\begin{tabular}{l|ccc|ccc}
\toprule[2pt]
\multicolumn{1}{c|}{\multirow{2}{*}{Method}} & \multicolumn{3}{c|}{ScanNet++~\cite{yeshwanth2023scannet++}} & \multicolumn{3}{c}{Replica~\cite{straub2019replica}} \\
\multicolumn{1}{c|}{}                         & PSNR$\uparrow$     & SSIM$\uparrow$   & LPIPS$\downarrow$    & PSNR$\uparrow$    & SSIM$\uparrow$    & LPIPS$\downarrow$   \\ \midrule
Mip-NeRF~\cite{barron2021mip}                & 19.58         &  0.755        &  0.389        & 18.12        & 0.707        & 0.391        \\
InfoNeRF~\cite{kim2022infonerf}     & 14.54         &  0.646        &  0.495        & 13.07        & 0.598     & 0.552        \\
DietNeRF~\cite{jain2021putting}     & 19.76         & 0.719       & 0.431        & 18.99       &0.676     &0.444    \\
FreeNeRF~\cite{yang2023freenerf}   &\cellcolor{yellow!25}20.17          &\cellcolor{yellow!25}0.756          & \cellcolor{yellow!25}0.368         &\cellcolor{yellow!25}20.99         &\cellcolor{yellow!25}0.765    &0.324         \\
Mip-NeRF$^*$~\cite{yu2022monosdf}                           &\cellcolor{orange!25}21.17          & \cellcolor{orange!25}0.779         &  0.395        &\cellcolor{orange!25}21.37         &\cellcolor{orange!25}0.785         &\cellcolor{yellow!25}0.318         \\ 
DNGaussian~\cite{li2024dngaussian} & 19.01 & 0.754 & \cellcolor{orange!25}0.367 & 17.63 & 0.718  & 0.435 \\
FSGS~\cite{zhu2025fsgs} & 17.95 & 0.730 & 0.373 & 20.22 & 0.760 & \cellcolor{orange!25}0.304 \\
\midrule[1.pt]
S$^3$NeRF (Ours)     & \cellcolor{red!25}{22.21}     &  \cellcolor{red!25}{0.787}   &  \cellcolor{red!25}{0.364} &    \cellcolor{red!25}{22.54}   & \cellcolor{red!25}{0.800}   &  \cellcolor{red!25}{0.287}        \\ 
\bottomrule[2pt]
\end{tabular}
}\vspace{-3mm}
\caption{Comparisons with other methods on ScanNet$++$ and Replica datasets. Mip-NeRF$^*$ refers to our baseline that utilizes estimated monocular depth for regularization. 
}\label{tab:sota}\vspace{-3mm}
\end{table}

\subsection{Comparison with Current Works}
\vspace{-1mm}
\noindent \textbf{Comparisons on Replica and ScanNet++. }We compare our S$^3$NeRF with several approaches on these two datasets, most of which target learning a NeRF from sparse inputs. Note that Mip-NeRF$^*$ applies the monocular depth regularization from MonoSDF~\cite{yu2022monosdf} which focuses on the sparse-input surface reconstruction. 
The quantitative results are shown in Tab.~\ref{tab:sota}. The results demonstrate that our S$^3$NeRF achieves the highest performance, outperforming FreeNeRF by more than 1.5 PSNR on both datasets. 
Fig.~\ref{fig:sota} illustrates that S$^3$NeRF shows better global structures and finer details compared with other works. For example, in the first row of examples in Fig.~\ref{fig:sota}, our S$^3$NeRF not only keeps more accurate details of the mouse, but also exhibits significantly fewer artifacts. Concerning the fourth row of examples, S$^3$NeRF shows fewer artifacts compared with InfoNeRF and FreeNeRF, and more accurate colors compared to DietNeRF. 
Our method also shows superior performance over the recent works of sparse-input 3D Gaussian Splatting (3DGS)~\cite{li2024dngaussian,zhu2025fsgs}. Note that 3DGS-based methods are optimized with random point cloud initialization, due to failures of applying COLMAP on these benchmarks resulting from less overlap across sparse input views. 

\begin{table}[t]
\Large
\centering
\resizebox{1.\linewidth}{!}{
\begin{tabular}{l|ccc|ccc}
\toprule[2pt]
\multirow{2}{*}{Method} & \multicolumn{3}{c|}{LLFF~\cite{mildenhall2019local}}   & \multicolumn{3}{c}{DTU~\cite{jensen2014large}}    \\
& \multicolumn{1}{c}{PSNR$\uparrow$} & \multicolumn{1}{c}{SSIM$\uparrow$} &\multicolumn{1}{c|}{LPIPS$\downarrow$} & \multicolumn{1}{c}{PSNR$\uparrow$} & \multicolumn{1}{c}{SSIM$\uparrow$} & \multicolumn{1}{c}{LPIPS$\downarrow$}\\ \midrule[1.2pt]
PixelNeRF ft~\cite{yu2021pixelnerf}      &            16.17              &    0.438         &0.512              &   18.95                       &   0.710  &0.269                     \\
MVSNeRF ft~\cite{chen2021mvsnerf}         &       17.88                   &    0.584            &\cellcolor{yellow!25}0.327           &   18.54     &      0.769       & 0.197            \\ 
RegNeRF~\cite{niemeyer2022regnerf}      & \multicolumn{1}{c}{19.08}     & \multicolumn{1}{c}{0.587}    & 0.336 & \multicolumn{1}{c}{18.89}     & \multicolumn{1}{c}{0.745}    & 0.190 \\
SimpleNeRF~\cite{somraj2023simplenerf}          & \cellcolor{yellow!25}19.24                         & \cellcolor{orange!25}0.623    & 0.375                      &   16.25                       &  0.751     & 0.249                   \\
GeCoNeRF~\cite{kwak2023geconerf}       &  18.77                        &  0.596               &0.338          &  -                       &   -       & -                \\
FreeNeRF~\cite{yang2023freenerf}     &  \cellcolor{orange!25}19.63                        &   \cellcolor{yellow!25}0.612                        & \cellcolor{orange!25}0.308&  \cellcolor{orange!25}19.92                        & \cellcolor{yellow!25}0.787  & \cellcolor{yellow!25}0.182 \\
SparseNeRF~\cite{wang2023sparsenerf}         &   \cellcolor{red!25}19.86                      &    \cellcolor{red!25}0.624      & 0.328                    & \cellcolor{yellow!25}19.55                         &  0.769     & 0.201                   \\
DNGaussian~\cite{li2024dngaussian}          &   19.12                       &    0.591              & \cellcolor{red!25}0.294         &    18.91                      &    \cellcolor{orange!25}0.790        & \cellcolor{orange!25}0.176              \\ \midrule[1.2pt]
SE-NeRF~\cite{jung2023self}           & \multicolumn{1}{c}{18.10}     & \multicolumn{1}{c}{0.540}   & 0.450  & \multicolumn{1}{c}{-}     & \multicolumn{1}{c}{-}  & -   \\
S$^3$NeRF (ours)       &  \cellcolor{red!25}19.86                        &   0.589             & 0.351           &    \cellcolor{red!25}21.09                      &  \cellcolor{red!25}0.835      & \cellcolor{red!25}0.147                  \\ \bottomrule[2pt]
\end{tabular}
}\vspace{-3mm}
\caption{Comparisons with other methods on LLFF and DTU datasets. Methods in the last block are based on self-improvement. }
\vspace{-5mm}
\label{tab:sota2}
\end{table}

\begin{figure}[t]
  \centering
  \vspace{-2mm}
  \includegraphics[width=0.98\linewidth]{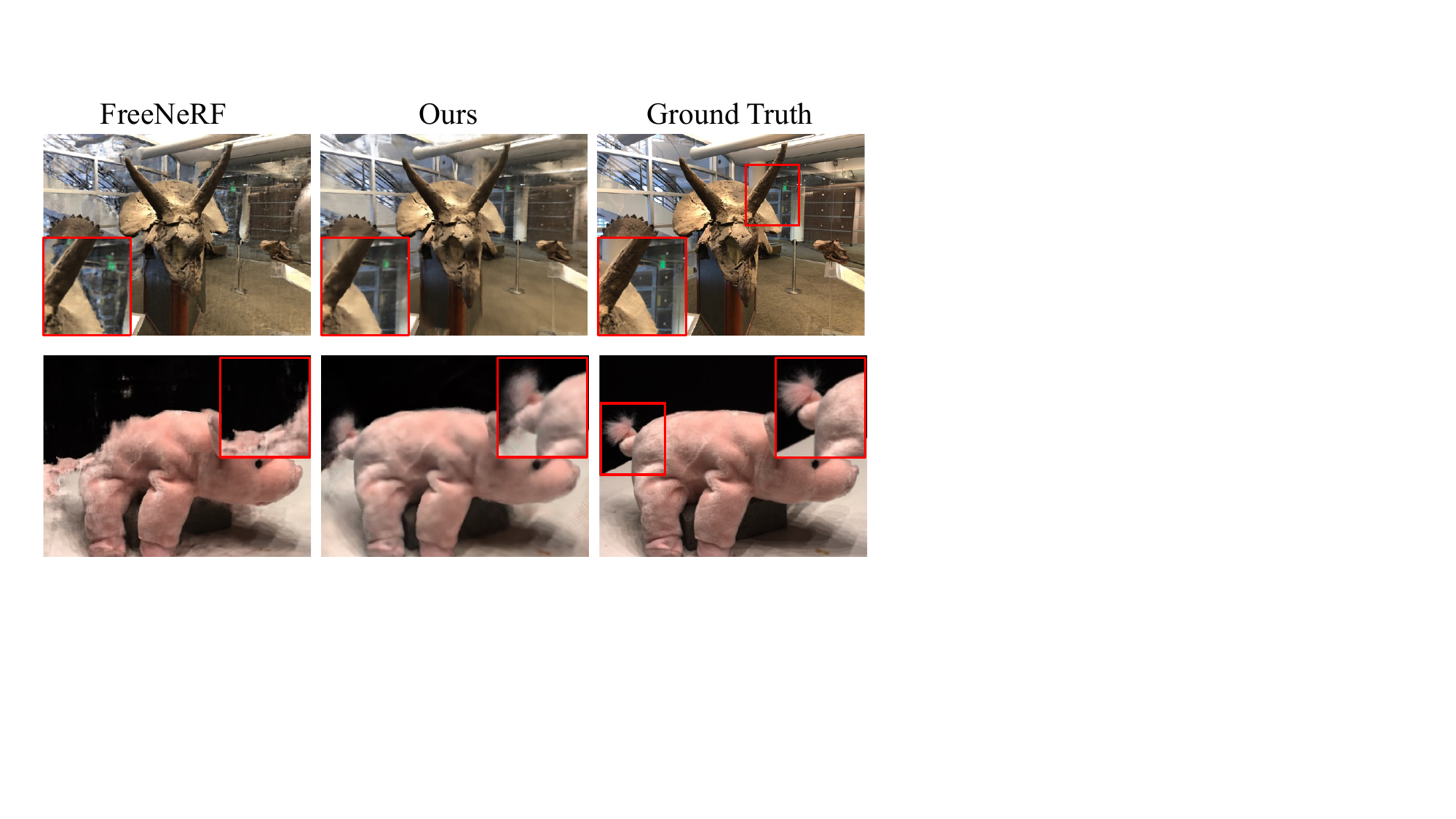}
  \vspace{-3mm}
  \caption{Qualitative comparisons on LLFF and DTU datasets. Our method preserves better structural integrity.  
  }\vspace{-5mm}
  \label{fig:sota_dtu}
\end{figure}

\noindent \textbf{Comparisons on LLFF and DTU. }We also compare our method with current approaches on widely used benchmarks. To get the semantic labels, we simply combine SAM~\cite{kirillov2023segment} and DINO~\cite{oquab2023dinov2} to get semantic labels of source views, which is detailed in the supplement. 
Most scenes in these benchmarks exhibit limited semantic differences, 
while Tab.~\ref{tab:sota2} surprisingly shows that our method achieves competitive performance. Fig.~\ref{fig:sota_dtu} qualitatively shows that our method can keep better global structure. Notably, our method outperforms another self-improved method SE-NeRF~\cite{jung2023self} by over 1.0 PSNR on LLFF, which utilizes rendered RGB from novel views for supervision, validating the superiority of the rendered semantics.

\vspace{-2mm}
\section{Conclusion}\vspace{-2mm}
This paper introduces an important observation that rendered semantics from dense novel views is a more effective form of augmented data than rendered RGB, based on which we 
propose a self-improved S$^3$NeRF to boost the sparse-input NeRF's performance. S$^3$NeRF encompasses supervision-level and feature-level guidance. 
The supervision-level guidance decides the validity of the rendered semantic labels by a Bi-Directional Verification module. The feature-level guidance further exploits the rendered semantic labels with a learnable codebook.  
We also introduce an additional indoor benchmark with as few as 6 input images. Experiments validate our approach and it achieves competitive performance compared with current works.

{
    \small
    \bibliographystyle{ieeenat_fullname}
    \bibliography{main}
}

\clearpage
\setcounter{page}{1}
\maketitlesupplementary

\renewcommand{\thefigure}{A\arabic{figure}}
\renewcommand{\thetable}{A\arabic{table}}
\renewcommand{\thesection}{\Alph{section}}
\setcounter{figure}{0}
\setcounter{table}{0}
\setcounter{section}{0}

\section{Implementation Details}
\noindent \textbf{Baseline. }We use the Mip-NeRF~\cite{barron2021mip} along with the estimated monocular depth for regularization~\cite{yu2022monosdf} as the baseline. The baseline serves as the teacher NeRF in the proposed S$^3$NeRF self-improved pipeline. We use MiDaS~\cite{ranftl2020towards} as the monocular depth estimator for each image of the sparse inputs. Besides the training objective of Eq.~3 in the main paper, we additionally sample a patch, and regularize its depth with the monocular estimated depth: 
\begin{equation}
    \mathcal{L}_{\rm{mono}} = \frac{1}{\left| \mathbf{r} \right|}  \sum_{\mathcal{P} \in \mathcal{R}} \sum_{\mathbf{r} \in \mathcal{P}}  \Vert (wD(\mathbf{r})+q-\bar{D}(\mathbf{r})) \Vert^2, 
\end{equation}
where $\mathcal P$ is the sampled patch from the sparse inputs, $w$ and $q$ are the scale and shift used to align the rendered depth $D$ and the estimated monocular depth $\bar{D}$. For more details, we refer readers to the paper of~\cite{yu2022monosdf}. Note that we do not claim this as our contribution, we just use it as a relatively strong baseline and prove the effectiveness of our method on it.

\noindent \textbf{Training. }To train the baseline (the teacher NeRF), for each iteration, we sample 2048 rays from sparse inputs for RGB supervision, and 1024 rays, i.e., 32$\times$16 patch from sparse inputs for monocular depth regularization. Hence, the batchsize of the baseline is 3072. After the baseline is trained, we train the semantic branch of Fig.~5(a) with 1024 rays with the semantic labels from sparse inputs, but fix the main branch. 

After the teacher NeRF is trained, we use it to render the outputs of novel views, consisting of rendered RGB, rendered depth, and rendered semantics. To train the student NeRF, for each iteration, except for the LLFF dataset, we sample 1024 rays from sparse inputs for both RGB supervision and semantics supervision, 1024 rays of 32$\times$16 patch for monocular depth regularization, and 1024 rays from novel view poses for the semantics supervision. 
For the LLFF dataset, we sample 3072 rays from sparse inputs for both RGB and semantics supervision, while others keep the same. 
The balancing term $\lambda_{\rm{sem}}$ in Eq.~3 is set to 0.1. 
Regarding the MLP structure with the proposed semantic-aware codebook, we set the number of codebook to 64, and the dimension $d$ of each word to 256. The attention of the querying process is multi-head, and we set the number of heads to 4. 

\begin{figure*}[t]
  \centering
  \includegraphics[width=0.85\linewidth]{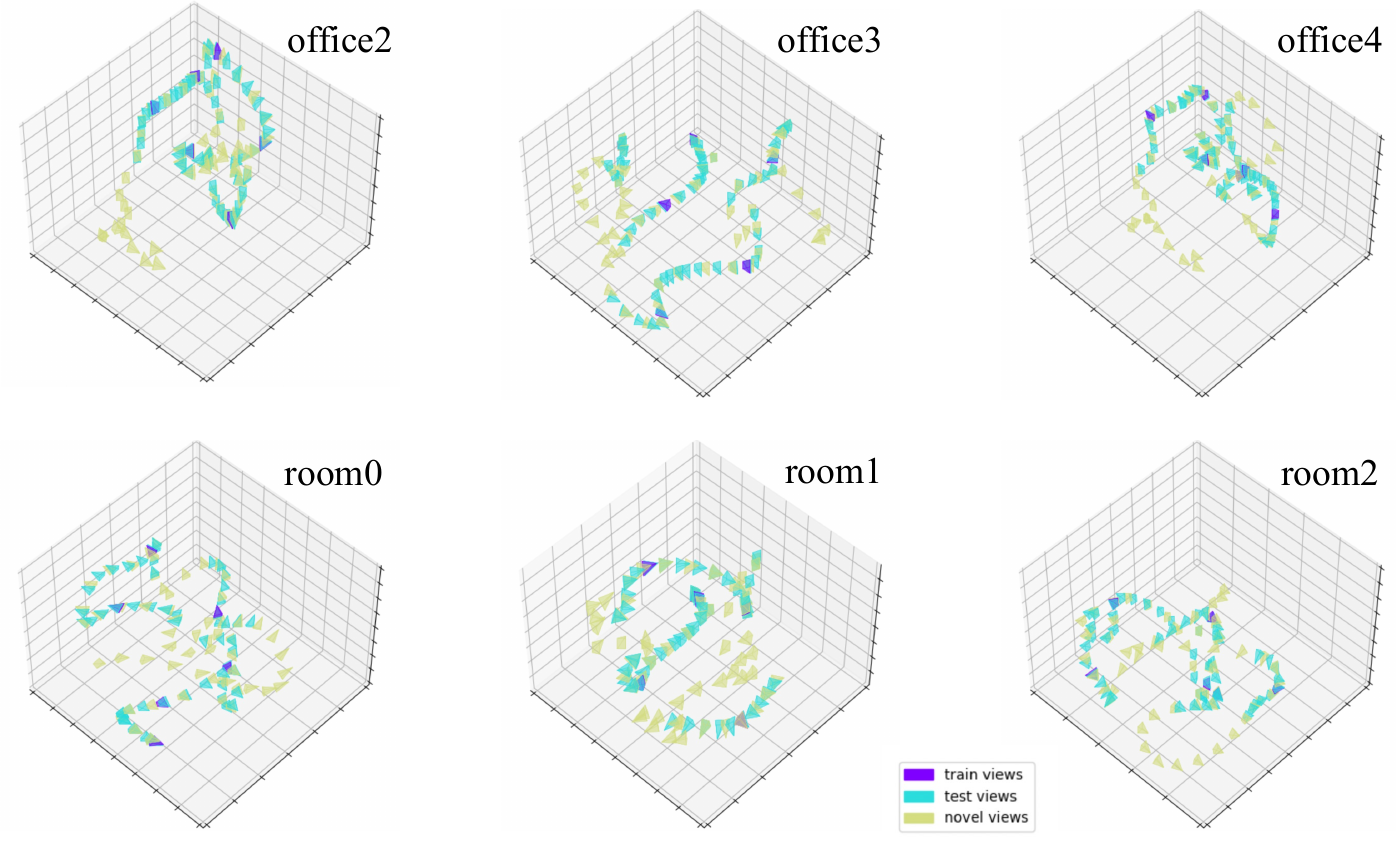}
  \caption{Illustration of camera poses of the Replica dataset, including training views of sparse inputs, test views, as well as novel views served as the augmented data. 
  }
  \label{fig:suppl_posevis}
\end{figure*}

\noindent \textbf{Novel view poses. }Our proposed S$^3$NeRF is a self-improved pipeline that leverages the rendered semantic guidance from novel view poses to improve the performance of the student NeRF. The teacher NeRF renders the outputs of novel view poses. While the poses of novel views can be arbitrary, for the sake of convenience, we utilize the poses of the camera trajectory already provided in each scene of the datasets as the novel view poses. 
We give an illustration of the camera poses of the Replica dataset in Fig.~\ref{fig:suppl_posevis}, regarding the positions and viewing directions. Note that, we do not leverage any forms of ground truth supervision from these novel views. The semantic guidance from these novel views is from the teacher NeRF. 

\noindent \textbf{Semantic label generation. }For the comparisons on DTU~\cite{jensen2014large} and LLFF~\cite{mildenhall2019local} datasets in the main paper, we combine SAM~\cite{kirillov2023segment} and DINOv2~\cite{oquab2023dinov2} to get semantic labels for source views. Concretely, for all source views, we use SAM to obtain all semantic masks and use DINOv2 to extract mask-wise features. Then, we apply K-Means on all these features and obtain a pseudo semantic label for each mask.

\section{Benchmark}
In the paper, we introduce a benchmark that is based on the ScanNet$++$~\cite{yeshwanth2023scannet++} and Replica~\cite{straub2019replica} datasets. The proposed benchmark is more challenging in its ``inside-out'' viewing direction as well as its limited input views of as few as 6. The selected 6 input views are sufficient to cover the entire scene, i.e., each part of the scene can be referred to a specific region in a certain input view.

The reasons for us to propose this benchmark are two-fold: \textbf{(1)} we want to validate our proposed method of leveraging rendered semantics in novel views. The proposed benchmark exhibits plentiful semantic information due to the indoor capturing, which facilitates our validation; \textbf{(2)} we want to study how well current sparse-input works perform on this more challenging less-overlap setting, which is of high-demand in real-world capture, but overlooked by current benchmarks.

 We select 4 scenes from the ScanNet$++$ dataset, and 6 scenes from the Replica dataset. The $\rm{ID}$ of the 4 scenes of the ScanNet dataset are: 8a20d62ac0, 94ee15e8ba, a29cccc784 and 7831862f02. The selected 6 scenes of the Replica dataset are: office2, office3, office4, room0, room1 and room2. The selected sparse inputs for all scenes are presented in Fig.~\ref{fig:dataset}. 

\noindent \textbf{Concerns about Camera Poses. }The less overlap across input views may sometimes lead to failures of COLMAP in estimating the camera poses. However, this should not be the concern of our proposed benchmark. We focus on scene modeling with given camera poses, following most of the sparse-input radiance fields reconstruction settings. 
This is a meaningful research setting. 
Currently, most real-world radiance fields reconstruction practices require given camera poses (regardless of the method used, e.g, COLMAP, DUSt3R~\cite{wang2024dust3r}, etc.) and ultimately train a radiance field. Even when the camera pose is accurate, results of the learned radiance fields are far from satisfactory with sparse inputs. 
Therefore, most current sparse-input research setting aim at addressing this difficulty by using groundtruth camera poses.

\section{Additional Quantitative Results}

\subsection{Per-Scene Performance}
We show the per-scene performance on Replica and ScanNet$++$ in Tab.~\ref{tab:suppl_per_scene}. The results show that, our method achieves competitive performance regarding three evaluation metrics in most scenes. When it comes to the average performance on each dataset, our method achieves highest performance compared to other works.

\begin{table*}[t]
\centering
\scriptsize
\begin{tabular}{c|ccccc|ccccccc}
\toprule
\multirow{2}{*}{}           & \multicolumn{5}{c|}{ScanNet$++$~\cite{yeshwanth2023scannet++}}                                                     & \multicolumn{7}{c}{Replica~\cite{straub2019replica}}                                                                                          \\
                            & \textit{a2ccc}          & \textit{8a20d}          & \textit{94ee1}          & \textit{78318}          & \textit{avg}          & \textit{office2}        & \textit{office3}        & \textit{office4}        & \textit{room0}          & \textit{room1}          & \textit{room2}          & \textit{avg}           \\ \midrule
\multirow{2}{*}{Mip-NeRF}   & 18.28          & 23.48          & 16.93          & 19.63          & 19.58          & 17.43          & 19.04          & 19.08          & 17.46          & 16.57          & 19.16          & 18.12          \\
                            & 0.759          & 0.799          & 0.725          & 0.735          & 0.755          & 0.539          & 0.685          & 0.727          & 0.762          & 0.721          & 0.808          & 0.707          \\
{~\cite{barron2021mip}}                     & 0.351          & 0.321          & 0.431          & 0.451          & 0.389          & 0.486          & 0.421          & 0.393          & 0.342          & 0.386          & 0.317          & 0.391          \\ \midrule
\multirow{2}{*}{InfoNeRF}   & 13.90          & 17.69          & 14.34          & 12.21          & 14.54          & 13.66          & 12.53          & 11.51          & 12.58          & 14.11          & 14.00          & 13.07          \\
                            & 0.662          & 0.691          & 0.627          & 0.605          & 0.646          & 0.463          & 0.545          & 0.592          & 0.618          & 0.689          & 0.678          & 0.598          \\
{~\cite{kim2022infonerf}}     & 0.468          & 0.437          & 0.516          & 0.558          & 0.495          & 0.612          & 0.623          & 0.624          & 0.542          & 0.435          & 0.477          & 0.552          \\ \midrule
\multirow{2}{*}{DietNeRF}   & 20.67          & 23.00          & 15.34          & 20.02          & 19.76          & 19.12          & 19.35          & 18.97          & 19.84          & 17.18          & 19.46          & 18.99          \\
                            & 0.751          & 0.776          & 0.627          & 0.725          & 0.719          & 0.612          & 0.695          & 0.419          & 0.783          & 0.749          & 0.797          & 0.676          \\
{~\cite{jain2021putting}}  & 0.385          & 0.363          & 0.516          & 0.459          & 0.431          & 0.458          & 0.417          & 0.721          & 0.34           & 0.386          & 0.343          & 0.444          \\ \midrule

\multirow{2}{*}{FreeNeRF}   & 19.93          & 22.37          & 19.42          & 18.94          & 20.17          & 20.89          & 21.06          & 20.25          & 22.55          & 19.69          & 21.43          & 20.99          \\
                            & 0.759          & 0.791          & 0.762          & 0.711          & 0.756          & 0.688          & 0.735          & 0.750          & 0.831          & 0.781          & 0.807          & 0.765          \\
{~\cite{yang2023freenerf}}                     & \textbf{0.307}          & 0.299          & 0.417          & 0.449          & 0.368          & 0.359          & 0.340          & 0.364          & 0.234          & 0.325          & 0.321          & 0.324          \\ \midrule

\multirow{2}{*}{DNGaussian}   & 19.10          & 21.21          & 17.55          & 18.20          & 19.01          & 22.68          & 18.40          & 12.31          & 12.60          & 18.87          & 20.91          & 17.63          \\
                            & 0.765          & 0.781          & 0.743          & 0.730          & 0.755          & \textbf{0.843}          & 0.789          & 0.644          & 0.534          & 0.708          & 0.790          & 0.718          \\
{~\cite{li2024dngaussian}}                     & 0.343          &\textbf{0.292}         & 0.382         & 0.450          & 0.367          & \textbf{0.233}          & 0.291          &0.628          & 0.722          & 0.397          & 0.338          & 0.435          \\ \midrule

\multirow{2}{*}{FSGS}   & 19.19          & 18.98          & 15.77          & 17.87          & 17.95          & 20.70          & 20.26          & 21.62          & 19.65          & 19.23          & 19.89          & 20.22          \\
                            & 0.760          & 0.735          & 0.719          & 0.708          & 0.730          & 0.802          & \textbf{0.790}         & \textbf{0.825}          & 0.654          & 0.712          & 0.779          & 0.760          \\
{~\cite{zhu2025fsgs}}                     & 0.321          & 0.316         & 0.415          & \textbf{0.442}          & 0.373          & 0.266          & \textbf{0.255}          & \textbf{0.271}          & 0.315          & 0.374          & 0.342          & 0.304          \\ \midrule

\multirow{2}{*}{Mip-NeRF$^*$}   & 20.68          & 24.66          & 18.30          & 21.04          & 21.17          & 21.77          & 22.43          & 21.93          & 22.05          & 18.80           & 21.26          & 21.37          \\
                            & 0.794          & 0.807          & 0.760          & \textbf{0.757}          & 0.779          & 0.721          & 0.747          & 0.791          & 0.836          & 0.789          & 0.824          & 0.785          \\
{~\cite{yu2022monosdf}}                     & 0.342          & 0.399          & 0.394          & 0.446          & 0.395          & 0.345          & 0.347          & 0.334          & 0.258          & 0.314          & 0.310          & 0.318          \\ \midrule
\multirow{2}{*}{S$^3$NeRF} & \textbf{21.81} & \textbf{25.60} & \textbf{20.05} & \textbf{21.36} & \textbf{22.21} & \textbf{22.79} & \textbf{23.83} & \textbf{23.08} & \textbf{24.01} & \textbf{19.66} & \textbf{21.87} & \textbf{22.54} \\
                            & \textbf{0.801} & \textbf{0.811} & \textbf{0.784} & {0.753} & \textbf{0.787} & {0.728} & {0.773} & {0.801} & \textbf{0.862} & \textbf{0.808} & \textbf{0.825} & \textbf{0.800} \\
{(Ours)}                     & {0.324} & {0.330} & \textbf{0.357} & {0.444} & \textbf{0.364} & {0.326} & {0.309} & {0.301} & \textbf{0.213} & \textbf{0.277} & \textbf{0.293} & \textbf{0.287} \\ \bottomrule
\end{tabular}
\caption{Per-scene performance of various models on the ScanNet++ and Replica datasets. For each method, the three rows represent PSNR, SSIM, and LPIPS, respectively. \textit{avg} refers the the average performance across all scenes in each dataset. }\label{tab:suppl_per_scene}\vspace{-3mm}
\end{table*}

\begin{table*}[t]
\centering
\scriptsize
\begin{tabular}{lccc}
\toprule
 &PSNR$\uparrow$  &SSIM$\uparrow$  &LPIPS$\downarrow$ \\ \midrule
 baseline&21.37  &0.785  &0.318  \\
 \quad w/ rendered RGB & 21.54&0.794 &0.320 \\
 \quad w/ rendered RGB (filtered) & 21.72&0.797 &0.312 \\
 \quad w/ rendered RGB (filtered) \& rendered semantics (filtered) & 21.82&0.796 &0.305 \\
 \quad w/ warped RGB (filtered) &21.36 & 0.788&0.315 \\
 \quad w/ warped RGB (filtered) \& rendered semantics (filtered) &21.40&0.787 &0.310 \\
 baseline w/ rendered semantics (filtered)&\textbf{22.34}  &\textbf{0.798}  &\textbf{0.291}  \\
  \bottomrule
\end{tabular}
\caption{Comparisons between using RGB and semantics of novel views as the augmented data on the Replica dataset. `w/ rendered RGB' utilizes the images rendered from novel views as the augmented data for RGB supervision. `w/ warped RGB' warps the RGB values from source views of sparse inputs to the novel views by projection, and the RGB values are used for the supervision on novel views. `filtered' refers to filtering the invalid pixels, decided by the validity map from the proposed supervision-level guidance. The last row refers to model with the supervision-level guidance proposed in the paper. 
}\label{tab:suppl_rgb}\vspace{-3mm}
\end{table*}

\subsection{Augmented RGB vs. Semantics}
In this work, we propose leveraging semantic guidance rendered from the teacher NeRF to enhance the training of the student NeRF. To demonstrate the effectiveness of rendered semantics, we compare augmented data from novel views in the form of rendered semantics with rendered RGB.

Tab.~\ref{tab:suppl_rgb} presents the utilization of augmented data from novel views for RGB supervision. The results indicate that applying the rendered RGB values alone yields only a marginal improvement and even leads to a drop in LPIPS performance. This suggests that the rendered RGB values in novel views might contain certain levels of noise, particularly for views that deviate significantly from the given training views, as illustrated in Fig.~\ref{fig:suppl_posevis}.

Our proposed supervision-level guidance incorporates a Bi-Directional Verification module to determine the validity of each pixel in the novel views. We use the validity map to filter out RGB values identified as invalid, denoted as `filtered' in Tab.~\ref{tab:suppl_rgb}. With the validity map, the performance improves, for instance, the LPIPS improves from 0.320 to 0.312. Additionally, combining the rendered RGB with rendered semantics in novel views shows negligible improvement. These results demonstrate that relying solely on rendered semantics determined by the validity map achieves the highest performance, while rendered RGB values may hinder the performance of rendered semantics.

Another approach for RGB supervision in novel views involves warping RGB values from source views. However, since the accuracy of warped values in novel views depends heavily on the rendered depth, we also employ the validity map to determine the validity of the warped values, denoted as `w/ warped RGB (filtered)'. The relevant results in Tab.~\ref{tab:suppl_rgb} do not exhibit significant improvement over the baseline.

Therefore, in the sparse inputs setting, we observe that rendered semantics are more effective than rendered RGB as a form of augmented data.

\subsection{Parameter Analysis}
We conduct an analysis of the parameters associated with the codebook in the feature-level guidance, focusing on the scene a29cccc784 from the ScanNet$++$ dataset. Two essential parameters need to be determined for the codebook: the index of the MLP layer whose output is used for querying the codebook (referred to as `layer idx'), and the number of words, denoted as $K$, in the codebook.

Our MLP architecture comprises 8 fully-connected layers before the density prediction. To evaluate the impact of different `layer idx' values, ranging from 1 to 7 (with the index starting from 0), we ablate the model accordingly. For instance, `layer idx' of 1 signifies using the output from the 2nd MLP layer, while 7 represents the output from the 8th MLP layer. Tab.~\ref{tab:codebook_ablate} demonstrates that setting `layer idx' to various values generally improves performance compared to the model trained with the supervision-level guidance proposed in our paper. This suggests that incorporating semantic-relevant information from the codebook into the network can enhance performance in most scenarios. However, we observe a performance drop when `layer idx' is set to 3. This can be attributed to the concatenated feature from this layer, which includes positional-encoded coordinates, potentially making it challenging for subsequent MLP layers to decode the mixed features effectively.

\begin{table*}[t]
\centering
\scriptsize
\resizebox{0.65\textwidth}{!}{
\begin{tabular}{llllllllc}
\toprule
\multicolumn{1}{c}{layer idx} & \multicolumn{1}{c}{1} & \multicolumn{1}{c}{2}  & \multicolumn{1}{c}{3}  & \multicolumn{1}{c}{4}  & \multicolumn{1}{c}{5}  & \multicolumn{1}{c}{6}  & \multicolumn{1}{c}{\textbf{7}}   & w/ supervision-level \\ \midrule
\multicolumn{1}{c}{PSNR}                          & 21.36                 & 21.80                   & 17.50                   & 21.66                  & 21.81                  & 21.82                  & {21.81}                   & 20.71             \\
\multicolumn{1}{c}{SSIM}                          & 0.799                 & 0.802                  & 0.732                  & 0.804                  & 0.802                  & 0.801                  & {0.801}                   & 0.788             \\ \midrule
\midrule
\multicolumn{1}{c}{\textit{K}}         & \multicolumn{1}{c}{8} & \multicolumn{1}{c}{16} & \multicolumn{1}{c}{32} & \multicolumn{1}{c}{\textbf{64}} & \multicolumn{1}{c}{84} & \multicolumn{1}{c}{96} & \multicolumn{1}{c}{128} & w/ supervision-level \\ \midrule
\multicolumn{1}{c}{PSNR}                          & 21.87                 & 22.02                  & 21.91                  & {21.81}                  & 21.86                  & 21.86                  & 21.89                   & 20.71             \\
\multicolumn{1}{c}{SSIM}                          & 0.802                 & 0.803                  & 0.801                  & {0.801}                  & 0.801                  & 0.802                  & 0.801                   & 0.788             \\ \bottomrule
\end{tabular}}
\caption{Parameter analysis on the feature-level guidance, conducted on scene a29cccc784 of the ScanNet$++$ dataset. `layer idx' refers to the index of the MLP layer whose output is used to query the codebook, while $K$ represents the number of words in the codebook. 
The parameters we select in our final model are marked in bold. 
}\label{tab:codebook_ablate}\vspace{-3mm}
\end{table*}

Regarding the number of words in the codebook, we fix `layer idx' to 7 and conduct ablation experiments by varying $K$ to different numbers, as shown in Tab.~\ref{tab:codebook_ablate}. Surprisingly, we find that performance do not significantly fluctuate with different values of $K$, even when $K$ is set to 8. This indicates that a relatively small number of words is sufficient to encode the semantic-relevant information regarding the scene, including the correlations between semantics, colors, and densities.

Based on our analysis, the feature-level guidance is not overly reliant on specific parameter settings for the codebook. Therefore, we empirically set the two parameters to 7 and 64, respectively, for all experiments in our paper.

\begin{figure*}[th]
  \centering
  \vspace{-6mm}
  \includegraphics[width=0.9\linewidth]{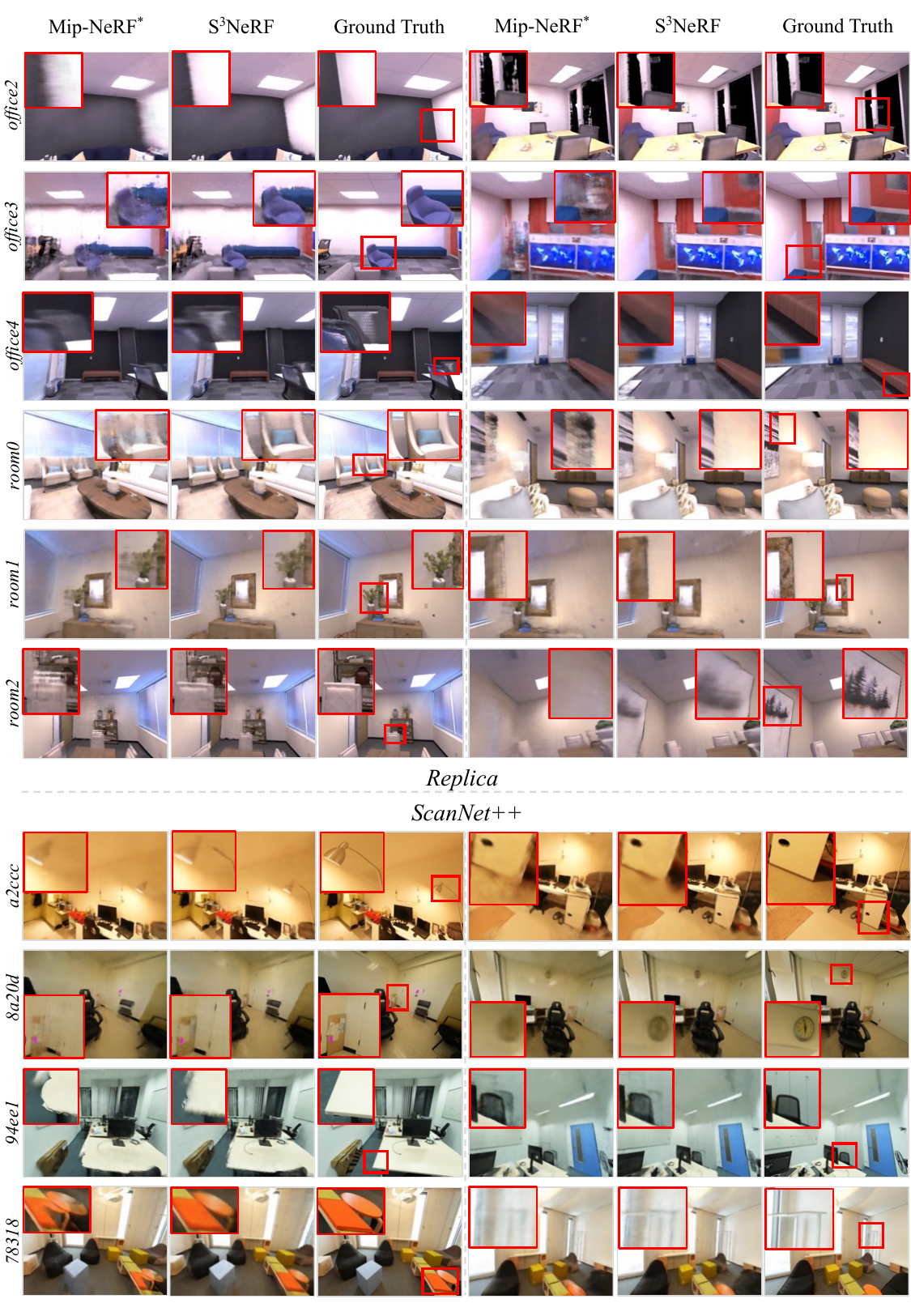}\vspace{-6mm}
  \caption{Qualitative comparisons between the proposed S$^3$NeRF and the baseline of Mip-NeRF$^*$ on the Replica and the ScanNet$++$ datasets. Mip-NeRF$^*$ leverages estimated monocular depth for regularization. 
  Each row in the figure displays rendered images from both methods, along with the corresponding ground truth images, for each scene. Two viewpoints are sampled for comparisons in each scene. }
  \label{fig:suppl_comparebase}
\end{figure*}

\section{Additional Qualitative Results}


\subsection{Comparison with the Baseline}
In this section, we compare our method S$^3$NeRF with the baseline Mip-NeRF$^*$, which is regularized by the estimated monocular depth. Fig.~\ref{fig:suppl_comparebase} illustrates qualitative comparisons for scenes from the Replica and ScanNet$++$ datasets. The results clearly demonstrate that S$^3$NeRF, with its rendered semantic guidance, significantly enhances visual quality compared to Mip-NeRF$^*$. S$^3$NeRF not only mitigates blurring and artifacts but also preserves sharper boundaries. These findings further substantiate the efficacy of the proposed supervision-level and feature-level guidance employed in S$^3$NeRF.

\subsection{Comparison with Current Methods}
In this section, we present additional qualitative results on two datasets, comparing our method S$^3$NeRF to InfoNeRF~\cite{kim2022infonerf}, DietNeRF~\cite{jain2021putting}, and FreeNeRF~\cite{yang2023freenerf}. Specifically, we showcase the results obtained on the ScanNet$++$ dataset in Fig.\ref{fig:suppl_vis_scannet}, and the results on the Replica dataset are depicted in Fig.\ref{fig:suppl_vis_replica}. These results provide further evidence of the superiority of our approach, both in terms of global structure and fine-grained details, when compared to other methods.

\begin{figure*}[t]
  \centering
  \vspace{-4mm}
  \includegraphics[width=0.9\linewidth]{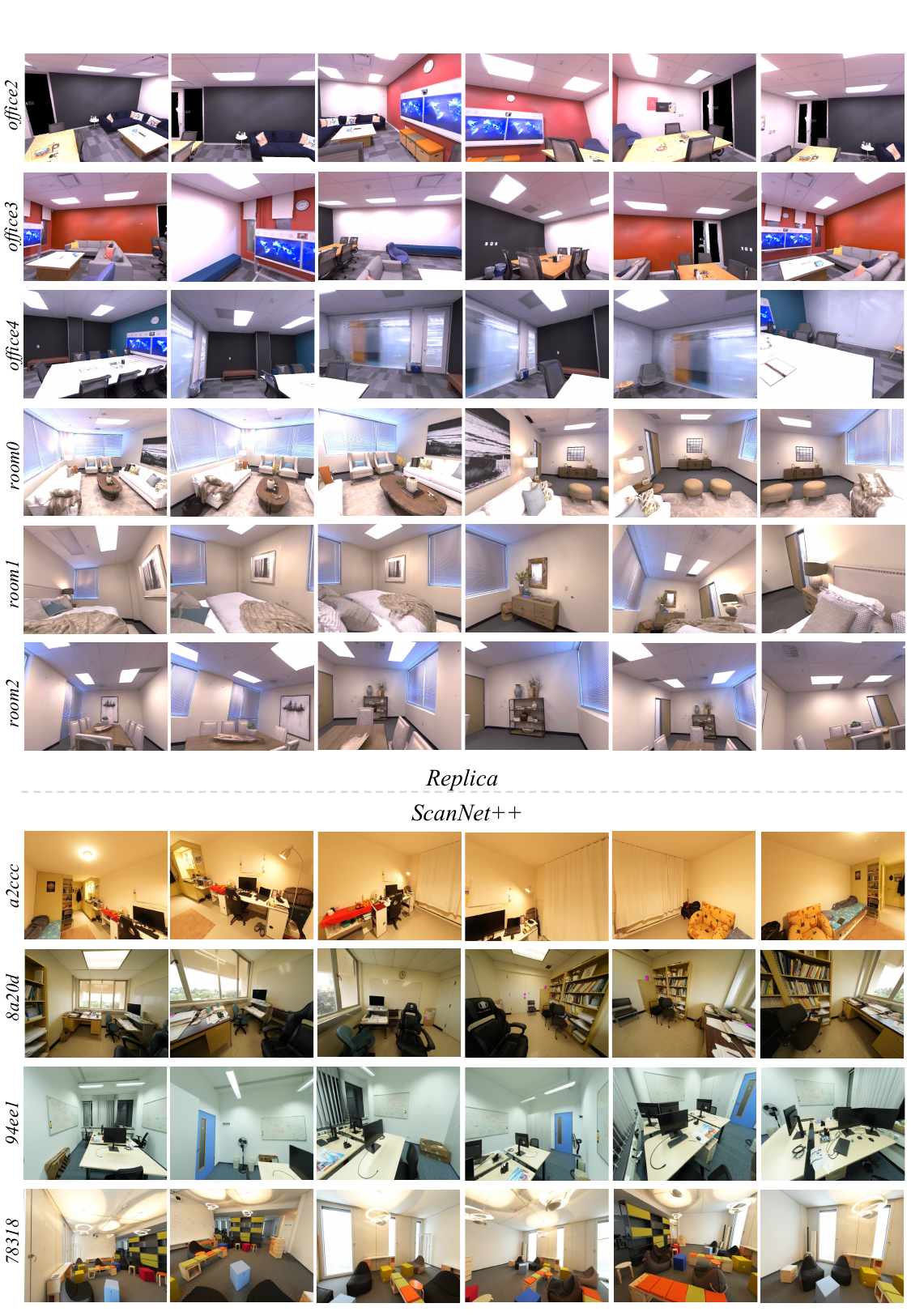}\vspace{-3mm}
  \caption{Illustration of sparse inputs consisting of 6 views. The sparse inputs provide comprehensive coverage of the entire scene.
  }\vspace{-3mm}
  \label{fig:dataset}
\end{figure*}

\begin{figure*}[t]
  \centering
  \includegraphics[width=0.99\linewidth]{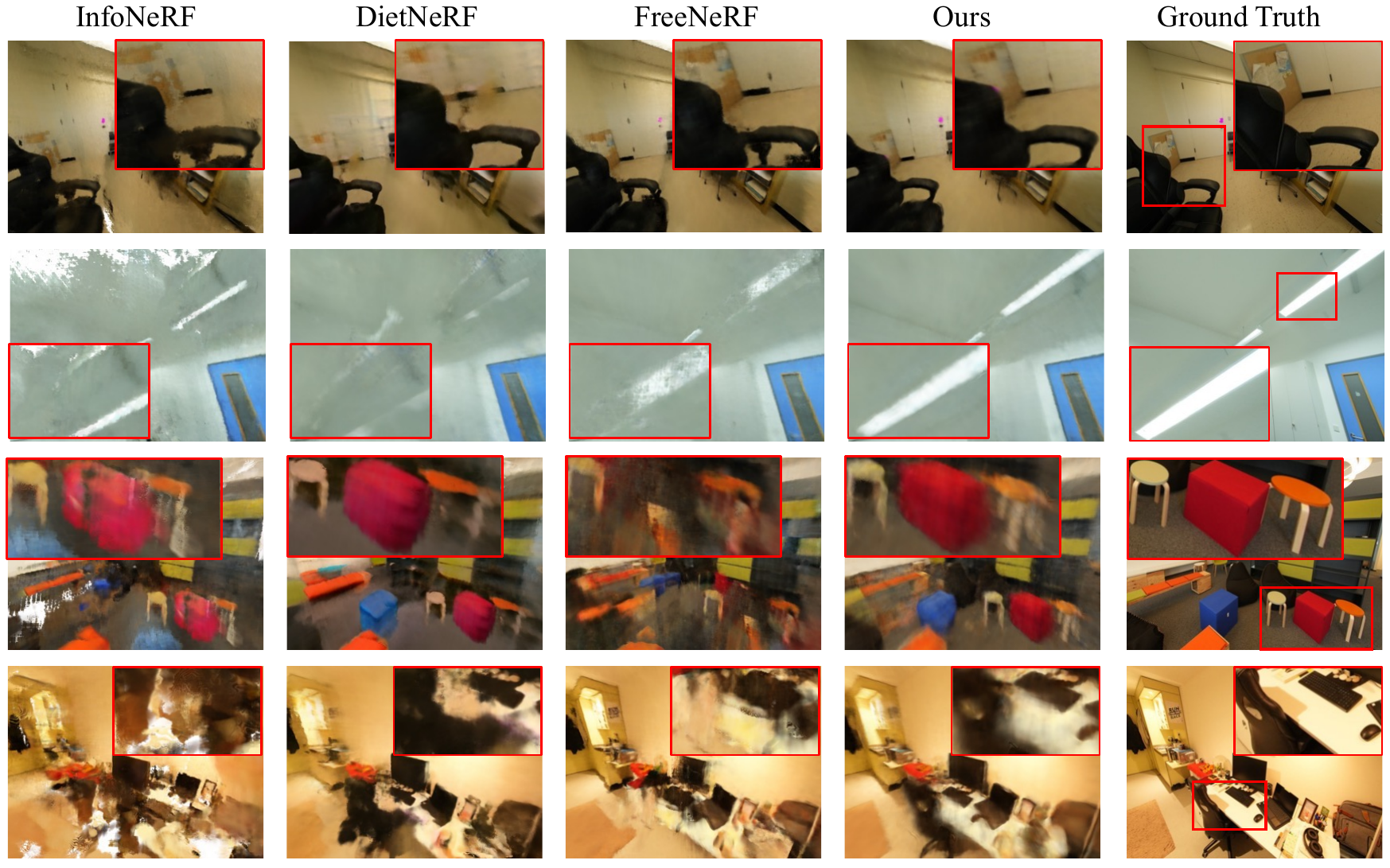}\vspace{-3mm}
  \caption{Qualitative comparisons with other works on the ScanNet$++$ dataset. }\vspace{-3mm}
  \label{fig:suppl_vis_scannet}
\end{figure*}

\begin{figure*}[t]
  \centering
  \includegraphics[width=0.99\linewidth]{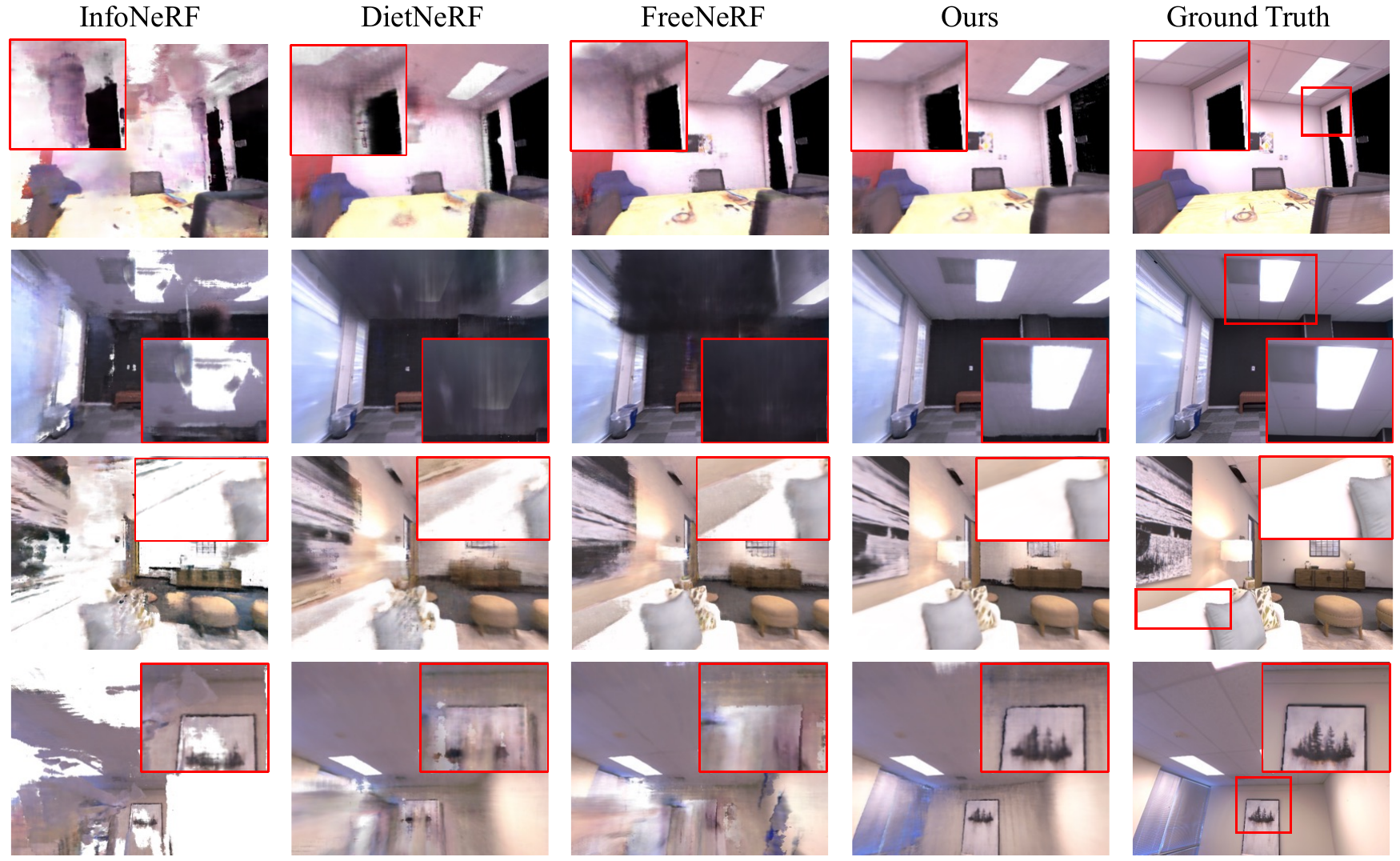}\vspace{-3mm}
  \caption{Qualitative comparisons with other works on the Replica dataset. }\vspace{-3mm}
  \label{fig:suppl_vis_replica}
\end{figure*}

\end{document}